\documentclass[twoside]{article}

\usepackage[preprint]{aistats2026}
\usepackage[round,sort]{natbib}

\usepackage[dvipsnames]{xcolor}
\usepackage{tikz}                  
\usepackage{macros}          
\usepackage{multicol}              
\usepackage[shortlabels]{enumitem} 
\usepackage{subfiles}              

\newenvironment{enum}{
    \begin{enumerate}[{\upshape (1)}, itemsep=-.33em]
}{
    \end{enumerate}
}

\usepackage[most]{tcolorbox}

\def\ourboxmargins{0.5em}
\newenvironment{blackblock}{
	\begin{tcolorbox}[
		enhanced,
		breakable,
		colback=black!10,
		leftrule=0.5mm,
		toprule=0pt,
		bottomrule=0pt,
		rightrule=0pt,
		arc=0pt,
		before skip=1em plus 2pt,
		after skip=1em plus 2pt,
		left=\ourboxmargins,
		right=\ourboxmargins,
		top=\ourboxmargins,
		bottom=\ourboxmargins,
		]
	}{
	\end{tcolorbox}
}

\begin{document}
    \allowdisplaybreaks

%

%

    \twocolumn[
        \runningtitle{Towards Blackwell Optimality: Bellman Optimality Is All You Can Get}
        \runningauthor{Victor Boone, Adrienne Tuynman}
        \aistatstitle{
            Towards Blackwell Optimality: 
            \\
            Bellman Optimality Is All You Can Get
        }
        \aistatsauthor{
            Victor Boone$^\dagger$
            \\ \texttt{victor.boone@irit.fr}
            \And 
            Adrienne Tuynman$^\ddagger$
            \\ \texttt{adrienne.tuynman@inria.fr}
        }

        \aistatsaddress{ 
            $^\dagger$Univ.~Grenoble Alpes, Inria, CNRS, Grenoble INP, LIG, 38000 Grenoble, France 
            \\
            $^\dagger$IRIT, Université de Toulouse, CNRS, Toulouse INP, Toulouse, France
            \\
            $^\ddagger$Univ.~Lille, Inria, CNRS, Centrale Lille, UMR 9189-CRIStAL, F-59000 Lille, France

        } 
    ]

    \begin{abstract}
    	Although average gain optimality is a commonly adopted performance measure in Markov Decision Processes (MDPs), it is often too asymptotic. 
        Further incorporating measures of immediate losses leads to the hierarchy of bias optimalities, all the way up to Blackwell optimality. 
        In this paper, we investigate the problem of identifying policies of such optimality orders. To that end, for each order, we construct a learning algorithm with vanishing probability of error. 
        Furthermore, we characterize the class of MDPs for which identification algorithms can stop in finite time. 
        That class corresponds to the MDPs with a unique Bellman optimal policy, and does not depend on the optimality order considered. 
        Lastly, we provide a tractable stopping rule that when coupled to our learning algorithm triggers in finite time whenever it is possible to do so. 
    \end{abstract}

    \section{INTRODUCTION}

What a simple learning task than the one of finding the optimal policy of a Markov decision process for which only a single reward --- or transition, is unknown. 
And yet, even if the uniqueness of the optimal policy is guaranteed beforehand, even if there is a single parameter to learn, even then, that learning task is surprisingly more delicate than it seems, especially when the optimality at stake is of higher order.
Because when one is trying to learn higher order optimalities, standard learning approaches and intuitions fail. 

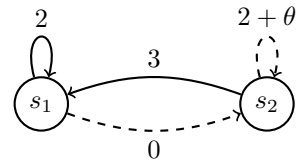
\begin{figure}[!ht]
	\centering
	\begin{tikzpicture}[node distance={30mm},thick,main/.style = {draw,circle}]
		\node[main] (1) {$\state_1$};
		\node[main] (2) [right of=1] {$\state_2$};
		\draw[->] (1) to [loop above,looseness=10]node{$2$} (1);
		\draw[->,dashed] (2) to [loop above,looseness=10]node{$2+\parameter$} (2);
		\draw[->] (2) to [out=160,in=20]node[midway, above,sloped,looseness = 15]{$3$} (1);
		\draw[->,dashed] (1) to [out=-20,in=-160]node[midway, below,sloped,looseness = 15]{$0$} (2);
	\end{tikzpicture}
	\caption{
        A class of deterministic transition Markov decision processes parameters by $\parameter \in (- \epsilon, \epsilon)$. 
    Choices of actions are represented by arrows that represent transitions.
    Labels are mean rewards.
    Rewards are reduced Gaussians, i.e., of the form $\NormalDistribution(-, 1)$. 
    }
	\label{figure_introductory_example}
\end{figure}

Let us consider the simple learning problem depicted in \Cref{figure_introductory_example}. 

In \Cref{figure_introductory_example}, we consider a class of deterministic Markov decision processes $\models \equiv \braces{\model_\parameter: \parameter \in (-\epsilon, \epsilon)}$ where the only thing to learn is the variable $\parameter$ in a neighborhood of $0$. 
In all cases, the full-line policy has an asymptotic reward of $2$, and when starting in state $\state_2$, an immediate reward of $3$.
Meanwhile, the dashed-line policy has an asymptotic reward of $2+\parameter$ and an immediate reward of $0$ when starting in $\state_1$. 
Clearly, the full-line policy is the best when $\parameter < 0$, while the dashed-lined one is better when $\parameter > 0$.
In case of the tiebreak $\parameter = 0$, we see that the full-line policy is strictly better than the other because of its immediate reward from $\state_2$ --- in formal terms, it is said bias-optimal. 
At $\parameter = 0$, we do not have both policies optimal from a higher order viewpoint. 
As such, the set of optimal policies is discontinuous at $\parameter = 0$.
Because no statistical test can ever determine that $\parameter = 0$, it is impossible to determine which policy is optimal, despite there always only being one.

So, is bias-optimality impossible to learn?

Despite its simplicity, \Cref{figure_introductory_example} provides a glimpse at the answer: 
if ``$\parameter = 0$'' is likely to hold under the current statistical data, the full-line policy is a better guess for bias-optimality than the dashed-line one. 
The idea is that if $\parameter \ne 0$, the learning agent should be able at some point to reject the possibility that ``$\parameter = 0$''.  
So, claiming that the full-line policy is bias-optimal --- when ``$\parameter = 0$'' is likely enough --- is the better temporary guess, although it can never be quantifiably certain. 

In this paper, we investigate the problem of learning high order optimalities in Markov decision processes and generalize the conclusions drawn from the example of \Cref{figure_introductory_example}.
We show that the unicity of the optimal policy is not determining for its learnability, and that having non unicity in lower orders can make the learning task impossible. Indeed, there exists one order of optimality that decides the learnability of all others: Bellman optimality.

    \section{PRELIMINARIES}
    
Let us start by giving a few common definitions and notations for our setting.

\subsection{MDPs and high order optimalities}

A Markov decision process is a tuple $\model \equiv (\Pairs, \nu, p)$, where $\Pairs \equiv \prod_{s \in \States} \{s\} \times \Actions(s)$ is the pair space, constructed as the dependent product of the state space $\States$ with the action space $\Actions$; the function $\nu : \Pairs \rightarrow \mathcal{P}([0,1])$ is the reward function, that maps state-action pairs to probability distributions over $[0, 1]$; the function $p : \Pairs \rightarrow \mathcal{P}(\States)$ is the transition kernel, that maps pairs to probability distributions over states.
The mean reward of the pair $\pair \in \pairs$ is the expected reward upon playing $\pair$ and is given by $\reward(\pair) := \integral x~\dd \rewarddistribution(\pair)(x)$. 
We finally define the distance between two models $M$ and $M'$ by 
\begin{equation*}
    \mdpdistance(M,M') 
    := 
    \max \braces*{
        \norm{r-r'}_\infty, \norm{p-p'}_\infty
    }
\end{equation*}

\subsubsection{Interacting with a MDP}

At time $t \ge 1$, the controller observes the current state $\State_t$, picks an action $\Action_t \in \actions(\State_t)$ and observes a reward $\Reward_t \sim \rewarddistribution(\State_t, \Action_t)$ and the next state $\State_{t+1} \sim \kernel(\State_t, \Action_t)$, both sampled independently from past observations.
The played pair is denoted $\Pair_t \equiv (\State_t, \Action_t)$.
The resulting interaction is summarized into a single vector $\History_t := (\State_1, \Action_1, \Reward_1, \State_2, \ldots, \State_t)$, called the history of observations. 
That vector lives in the history space, given by $\histories := \bigcup_{t = 1}^\infty \histories_t$ where $\histories_t := (\pairs \times [0, 1])^{t-1} \times \states$. 
Formally, the choice of action $\Action_t$ is $\sigma(\History_t)$-measurable and can be described as a kernel $\learner : \histories \to \actions$ from possible histories to actions, that we refer to as a \strong{sampling rule}.
Finally, the choice of a Markov decision process $\model$, of a sampling rule $\learner$ and of an initial state $\state_0 \in \states$ properly defines a probability space on the history space $\histories$, of which we write $\Pr_ {\state_0}^{\model, \learner}(-)$ and $\EE_{\state_0}^{\model, \learner}[-]$ the probability and expectation. 
Often, the Markov decision process is controlled by sampling rules that are independent of time, called \strong{policies}, written $\policy \in \policies := \states \to \actions$.

A few assumptions must be added to the Markov decision process to guarantee that learning objectives can be met. 
In some models, there are multiple groups of disconnected states: some states are not reachable from others. 
If high rewards are in those unreachable states, how is the agent supposed to learn them? 
Learning agents need to explore everywhere, and should not be punished for ending up trapped in low-reward states. 
To prevent this problem, we will exclusively consider models in which good states are always accessible. 

\begin{assumption}
    The model $\model$ is \strong{communicating}.
    That is, for every pair of states $s,s'$, there exists a deterministic stationary policy $\pi$ such that
    \begin{equation*}
        \exists n \ge 1,
        \quad
        \Pr^{M,\pi}_s(\State_n=s')
        > 0
        .
    \end{equation*}
\end{assumption}

Finally, in some cases, we will look closely at what we call \strong{unichain policies}. Those are policies $\pi$ such that states are either reachable from every other state (and thus are visited infinitely many times, whatever the starting state), or are not reachable from the first set of states (and are thus visited finitely many times, whatever the starting state).

\subsubsection{Gain, High Order Biases \& Optimalities}

Given a policy $\pi \in \Pi$, its \strong{gain} and \strong{bias} functions are:
\begin{equation*} 
\begin{aligned}
	\gain^\pi (s,M) 
    & := 
    \lim_{T \rightarrow +\infty} 
    \frac{1}{T} 
    \bE_\state^{M,\pi} \brackets*{
        \sum_{t=1}^T 
        R_t
    }
    \\
    \bias^\pi (s,M)
    & :=
    \Clim_{T \rightarrow +\infty} 
    \bE_s^{M,\pi} \brackets*{
        \sum_{t=1}^T 
        (R_t - g^\pi (S_t,M))
    }
    .
\end{aligned}
\end{equation*}
That is, the gain is the asymptotic average amount of reward that the policy collects, while the bias is the amount of immediate reward gathered compared to the gain. 
These quantities will also be written as $\nbias_{-1} \equiv \gain$ and $\nbias_0 \equiv \bias$.
By convention, we let $\nbias_{-2} = 0$. 
Beyond the gain and the bias, the $n$-th order bias $\nbias^\pi_n$ of $\pi$ (for $n \geq 1$) is the bias of $\pi$ for which the reward function $r(s,a)$ is changed to the previous bias value $- \nbias^\pi_{n-1} (s)$:
\begin{equation*}
    \nbias_n^\policy (\state; \model)
    :=
    - \Clim_{T \rightarrow +\infty}
    \EE_{\state}^{\model, \policy} \brackets*{
        \sum_{t=1}^T
        \nbias_{n-1}^{\policy}(\State_t; \model)  
    }
    .
\end{equation*}

The \strong{optimal gain} and \strong{bias} functions are defined in a nested fashion.
For $n = -2$, we set $\optpolicies_{-2}(\model) = \policies$ as the set of all policies.
For $n \geq -1$, we define the optimal bias $\nbias^\opt_n (\model)$ of order $n$ as well the set of optimal policies $\optpolicies_n(\model)$ of order $n$, also called \strong{$n$-optimal policies}, in a co-inductive way:
\begin{equation*}
\begin{cases}
    &
    \displaystyle
    \nbias^\opt_n (s,M) 
    := 
    \sup_{\pi \in \Pi^\star_{n-1}(M)} \nbias^\pi_n (s,M)
    \\
    &
    \displaystyle
    \optpolicies_n (\model)
    := \braces*{
        \policy \in \policies:
        \forall k \le n, \nbias^{\policy}_k (\model) = \nbias^{\opt}_k (\model)
    }
    .
\end{cases}
\end{equation*}
Note that $\Pi^\star_n(M) \subseteq \Pi^\star_{n-1}(\model)$. 
The set of \strong{Blackwell optimal policies} is the set of policies that are optimal at every order, i.e., $\Pi^\star_\infty (M) := \bigcap_{n=-1}^{+\infty} \Pi_n^\star(M)$. 
It is known that Blackwell optimal exist unconditionally, i.e., that $\Pi^\star_\infty (M) \neq \emptyset$, see \cite[Theorem~10.1.4]{puterman_markov_1994} for example.

Lastly the $n$-th order \strong{gaps} relative to a policy $\policy \in \policies$ are defined as
\begin{equation*}
    \Delta_n^{\pi}(s,a):= \nbias_n(s) +\nbias_{n-1}(s) -p(s,a) \nbias_n -r_n(s,a)
\end{equation*}
for $(s,a)$ a state action pair, where $r_n := \ind_{n=0} r$ is the $n$-th order reward function.
They lead to what we call high order Bellman optimal policies. 

\begin{definition}
\label{definition_bellman_optimal_high_order}
	A policy is \strong{$n$-Bellman optimal} if, for all $m\leq n$, we have for every state action pair $\pair \in \pairs$,
    \begin{equation*}
        [\forall \ell<m, \gaps_\ell^\pi(\pair)=0] 
        \implies 
        \gaps^\pi_m(\pair)\geq 0
    \end{equation*}
\end{definition}

As a matter of fact, \Cref{definition_bellman_optimal_high_order} states that the policy satisfies the first $n+2$ nested Bellman optimality equations \cite[§10.2]{puterman_markov_1994}.
In other words, a policy is $n$-Bellman optimal if, to every order up to $n$, it beats the other actions that were optimal until then.
When we say that policy $\policy$ is \strong{Bellman optimal} without specifying the order, we mean that it is $0$-Bellman optimal, and we write $\policy \in \bellmanpolicies(\model)$. 
This notion of Bellman optimality coincides with the one motivated by \cite{gast_testing_2023}. 
As we will see further down, those Bellman optimal policies have special connections to optimal policies of order $n$.

\subsection{Fixed-Confidence Identification and \texorpdfstring{$\delta$}{delta}-Probably Correct Algorithms}
\label{section_identification_taks}

In identification, the underlying Markov decision process is unknown and the objective is to find an optimal policy as quickly as possible.
As such, the learning algorithm explores the hidden Markov decision process by playing actions.
Doing so, it gathers information from which a stream of potential optimal policies is produced. 
Formally, an identification algorithm consists in a pair of 
\begin{enum}
    \item 
        A \strong{sampling rule} $\learner : \histories \to \actions$ that decides how the Markov decision process is navigated;
    \item 
        A \strong{recommendation rule} $\recommendation : \histories \to \policies$ that emits a stream of potential optimal policies.
\end{enum}
The recommended policy at time $t \ge 1$, denoted $\recommendation_t$, is the policy that the algorithm believes to be the best guess as an element of $\optpolicies(\model)$. 

From there, there are two main ways to appreciate the learning performance of a learning agent.

The more straight-forward approach is to minimize the probability of error $\Pr^{\model, \learner}(\recommendation_t \notin \optpolicies(\model))$ as $t$ grows. 
At the very least, one expects that this probability vanishes asymptotically, leading to the definition of consistency from \cite{kaufmann_complexity_2016}, see \Cref{definition_consistency} below. 

\begin{definition}[Consistency]
\label{definition_consistency}
    An identification algorithm $(\learner, \recommendation)$ is said \strong{consistent} with respect to a class of models $\models$ and an optimality map $\optpolicies : \models \to 2^\policies$ if 
    \begin{equation*}
        \forall \model \in \models,
        \quad
        \Pr^{\model, \learner}(\recommendation_t \notin \optpolicies(\model)) 
        = 
        \oh(1).
    \end{equation*}
\end{definition}

The natural performance criterion for consistent algorithms is the behaviour of the error probability $\Pr^{\model, \learner}(\recommendation_t \notin \optpolicies(\model))$ when $t \to \infty$. 

Beyond consistency alone, we are further interested in \emph{optimality certificates}. 
That is, in addition to the mere recommendation $\recommendation_t$, one requires the algorithm to estimate how likely $\recommendation_t$ is to be optimal.
In the literature, this learning task is typically formulated the other way around: one picks a confidence threshold $\delta > 0$ and looks for a stopping time $\tau_\delta$ at which the recommendation $\recommendation_{\tau_\delta}$ is correct with probability greater than $1 - \delta$. 
Such algorithms are said $\delta$-probably correct ($\delta$-PC, \cite{boone_identification_2023}), see \Cref{definition_delta_pc}.

\begin{definition}[Probable correctedness]
\label{definition_delta_pc}
    An identification algorithm $(\learner, \recommendation)$ together with a stopping time $\tau_\delta$ is said \strong{$\delta$-probably correct} ($\delta$-PC) with respect to a class of models $\models$ and an optimality map $\optpolicies: \models \to 2^\policies$ if 
    \begin{equation*}
        \forall \model \in \models, 
        \quad
        \Pr^{\model, \learner} \parens*{
            \recommendation_{\tau_\delta} \notin \optpolicies(\model)
        }
        \le
        \delta
        .
    \end{equation*}
\end{definition}

The natural performance criterion for $\delta$-PC algorithms is their \strong{sample complexity} $\EE^{\model, \learner}[\tau_\delta]$ (\cite{mannor_sample_2004,kaufmann_complexity_2016}), that measures the expected time before the algorithm stops. 

In this paper, we will be interested in consistent algorithms enriched with a $\delta$-PC property. 

\subsection{Related Works}

The problem of identifying optimal policies can be likened to best arm identification (\cite{mannor_sample_2004,audibert_best_2010}) in multi-armed bandits, that are \emph{de facto} state-less RL problems. 
The time complexity of best arm identification is thoroughly described by \cite{kaufmann_complexity_2016}, where the complexity is shown to stem from the proximity of the mean rewards. 
With bandits, however, an arm is optimal or it is not; hence, all optimality orders are absolutely equivalent, therefore occluding much of the nuances of the RL setting.

A worth-mentioning pool of works that sits closely to this work is perhaps the problem of the identification of gain optimal policies in average reward RL in the $(\epsilon, \delta)$-PAC setting (\cite{jin_towards_2021,wang_optimal_2024,zurek_span_2024}), that has seen significant advances in the recent years. 
In this setting, the goal is to find an $\epsilon$-gain optimal policy with probability $1 - \delta$ with as few samples of the underlying instance as possible. 
Despite the impossibility to estimate it (\cite{tuynman_finding_2024}), 
the bias of the optimal policy appears in the sample complexity bounds (\cite{zurek-spanagnostic2025,lee_near_2025}), highlighting the dependencies between the various optimality orders and the difficulties to determine them.

Beyond identification problems, other lines of work are focusing on the computation of Blackwell optimal policies.
Notably, \cite{grand-clementReducingBlackwellAverage2023,mukherjeeEfficientComputationBlackwell2025} have worked off the historical definition of \cite{blackwell_discrete_1962}, stating that a Blackwell optimal policy must be discount optimal for all discount factors beyond a threshold $\gamma_0 < 1$, providing new complexity guarantees regarding their computation. 
The value of the threshold is discontinuous in $\model$ however, thus complicating the approach when further incorporating noise and uncertainty. 

At the intersection of identification and Blackwell optimal policies, it has been shown recently that, in the space of Markov decision processes with deterministic transitions, the learnability of Blackwell optimal policies depends on a few criteria on the gain and bias optimal policies (\cite{boone_identification_2023}). 
More specifically, the uniqueness of the bias optimal policy, and the unichain character of all gain optimal policies, were shown to be necessary conditions for learnability.

\subsection{Contributions}

We start our paper by showing that Blackwell optimal policies can be computed even in the presence of noise. 
More precisely, in \Cref{sect:consistency}, we show that consistent algorithms exist. 
For that, we provide \texttt{HOPE}, standing for \texttt{H}igher \texttt{O}rder \texttt{P}olicy iteration \texttt{E}psilonized, that we show to be consistent for $\optpolicies_\infty$. 

We stress that a natural set of MDPs to focus on are those for which consistent algorithms can be stopped in finite time. 
Such instances are said \emph{non-degenerate}. 
Although non-degeneracy obviously depends on the order of optimality that we consider, we characterize non-degenerate MDPs at every order. 
To that end, we first prove in \Cref{sect:degen} that non-degenerate MDPs necessarily have a unique Bellman-optimal policy, generalizing the results of \cite{boone_identification_2023} to stochastic transition MDPs. 
Lastly, in \Cref{sect:stopping}, we provide a stopping rule for \texttt{HOPE}, therefore proving that MDPs with a unique Bellman optimal policy are non-degenerate. 
As a result, a MDP is non-degenerate if and only if it has a unique Bellman optimal policy, whatever the order of optimality that we consider.

    \section{CONSISTENT ALGORITHMS AND MODEL DEGENERACY}
    \label{sect:consistency}
    \label{section_consistency}

As shown in \Cref{figure_introductory_example}, simply returning the best policy of the empirical policy is not necessarily a good idea, as the probability of error is not necessarily asymptotically vanishing. 
This raises the following question: for some order $n$, is there an algorithm that is consistent for the optimality mapping $\Pi^\opt_n$?

\subsection{Existence of Consistent Algorithms}

We show that consistent algorithms exist indeed for every order. 
This means that there exist an algorithm $(\learner, \recommendation)$ with vanishing probability of error for any order $n \ge -1$ and model $\model$, i.e., $\Pr^{\model, \learner}(\recommendation_t \notin \optpolicies(\model)) \to 0$. 

For that, we build off Policy Iteration from \cite{puterman_markov_1994}.
The Policy Iteration algorithm maintains a policy that is improved gradually using a subroutine called Policy Improvement.
That subroutine guarantees that the aggregate bias vector $(\nbias_m^\policy)_{m \ge -1}$ is improved everytime the policy is updated. 
At a proof level, the algorithm progressively discovers the optimal biases $(\nbias_m^*)_{m \ge -1}$.
Once the algorithm is sure that $\nbias_m^\policy = \nbias_m^*$ for $m \le k$, it will restrict the Policy Improvement subroutine to select state-action pairs such that $\gaps^\policy_m (\state, \action) = 0$ for $m \le k-1$.
At that point, the policy is improved by looking for state-action pairs for which $\gaps^\policy_k (\state, \action) < 0$ or $\gaps^\policy_{k+1} (\state, \action) < 0$. 
The issue of replicating this algorithm in a learning setting is the constraint 
\begin{equation}
\label{equation_constraint_gap}
    \gaps_m^\policy (\state, \action) = 0,
    \quad \forall m \le k-1
\end{equation}
because the estimation of the biases $(\nbias_m^\policy)$, therefore of the gaps $\gaps_m^\policy$, is noisy. 

\paragraph{Dealing with noise in Policy Iteration.}
We thus construct Higher Order Policy Iteration (\texttt{HOPI}$(n,\varepsilon)$), \Cref{alg:HOPI} (described in detail in \Cref{app:consistency}), that, for each order, finds the set of policies that are \textit{nearly} Bellman optimal for that order.
It does so by relaxing the constraint \eqref{equation_constraint_gap} to
\begin{equation}
    \gaps_m^\policy (\state, \action) \le \epsilon,
    \quad \forall m \le k-1
    .
\end{equation}
In the example of Figure~\ref{figure_introductory_example} with $\theta=0$, even if $\theta$ is a bit overestimated, the full-line policy will remain near gain-optimal, and will thus be considered for bias optimality instead of being disqualified.

Running \texttt{HOPI} with a small enough $\varepsilon$ will return exactly the correct policies, even in the presence of noise --- provided that this noise is small enough.
Indeed, a small noise on the MDP translates to a small noise in the biases. We will show in \Cref{lemma_close_MDPs} that there exists a function $\psi: \RR_+ \to [0, \infty]$ with
\begin{equation*}
    \sup_{\policy \in \policies}
    \abs*{r_n+p \nbias_n^\pi-r'_n -p'{\nbias'}_n^\pi}
    \le
    \psi(\mdpdistance(\model, \model'))
\end{equation*}
such that $\psi(\epsilon) = \OH(\epsilon)$ when $\epsilon \to 0$. 

In \Cref{corollary_bissimulation}, we use that result to show that there exists a certain function $f$ such that, when $\mdpdistance(M,M')\leq f(\varepsilon,M)$, $\texttt{HOPI}(n,0,M)$ gives the exact same result as $\texttt{HOPI}(n,\varepsilon,M')$, because the sequence of policies of \texttt{HOPI}$(n, \epsilon, \model')$ is a valid sequence of policies for \texttt{HOPI}$(n, 0, \model)$. 

This is the basis for our Higher Order Policy iteration Epsilonized (\texttt{HOPE}) algorithm (\Cref{alg:HOPE}).

\begin{algorithm}
	\caption{Higher Order Policy iteration Epsilonized}
	\label{alg:HOPE}
	\begin{algorithmic}
        \REQUIRE Desired order of optimality $n \ge -1$. 
		\STATE $t\gets 0$
		\LOOP
		\STATE Construct empirical MDP $\hat{\model}_t$
		\STATE $\varepsilon \gets t^{-1/4}$
		\STATE $ \prod_{s} \actions_n(s) \gets \texttt{HOPI}_{n,\varepsilon}(\hat{\model}_t)$
		\STATE Recommend any policy in $\prod_{s} \actions_n(s)$
		\STATE Sample uniformly $a_t \in \actions (s_t)$, observe $r_t$ and $s_{t+1}$
		\STATE $t\gets t+1$
		\ENDLOOP
	\end{algorithmic}
\end{algorithm}

We will show in \Cref{app:consistency} that the choice of $\varepsilon_t :=t^{-1/4}$ is sufficient to get vanishing probability of error.

\begin{theorem}
\label{th_consistency}
    For all $n \ge -1$, \texttt{HOPE}$(n)$ (\Cref{alg:HOPE}) is consistent for $\optpolicies_n$.
\end{theorem}

\subsection{MDPs Learnable in Finite Time}

Since consistent algorithms exist, we have shown that against all odds, high order optimal policies can be found even in the presence of noise. 
However, even if the recommendation is correct \textit{at some point}, there might be no way of knowing when this happens. 
In \Cref{figure_introductory_example}, for instance, since the optimal policy of $M_\theta$ changes depending on whether $\theta \ge 0$ or $\theta <0$, it is not possible to reject $\theta <0$ in $\model_0$.
The problem consists in giving a finite stopping time, that is, being able to at some time $t$ give a certificate of optimality of the recommended policy. 
We define non-degenerate models as those in which, for some consistent algorithm, there exists a stopping time that is finite on the model.

\begin{definition}[Non-degenerate models]
	\label{definition_non_degenerate}
	A model $\model \in \models$ is said non-degenerate with respect to an optimality map $\optpolicies$ if there exists a consistent algorithm $(\learner, \recommendation)$ and a collection of stopping rules $(\tau_\delta)_{\delta > 0}$ such that, regardless of the confidence parameter $\delta > 0$, we have
	\begin{enum}
		\item $(\learner, \recommendation, \tau_\delta)$ is $\delta$-PC on $\models$ with respect to $\optpolicies$;
		\item $\tau_\delta$ is almost-surely finite in $\model$, i.e., we have $\Pr^{\model, \learner}(\tau_\delta < \infty) = 1$. 
	\end{enum}
\end{definition}

The question that we raise now is the following: which models are degenerate according to this definition?
On which models is the construction $\delta$-PC algorithms hopeless? 
This is the central question of our paper, and here is our final answer:

\begin{blackblock}
\begin{theorem}[Characterization of non-degeneracy]
	\label{theorem_characterization_non_degeneracy}
	Let $n \ge -1$.
	A Markov decision process $\model \in \models$ is non-degenerate w.r.t.~$\optpolicies_{n}(\model)$ if, and only if $\model$ has a unique Bellman optimal policy.
\end{theorem}
\end{blackblock}

It is well-known that once the Bellman optimal policy $\optpolicy$ is unique, then it is not only the unique bias optimal policy, but also the unique Blackwell optimal policy, i.e., that $\optpolicies_0 (\model) = \optpolicies_1 (\model) = \ldots = \optpolicies_\infty (\model) = \braces{\optpolicy}$, see \cite[Corollary 10.1.7]{puterman_markov_1994}.

Combined with this observation, \Cref{theorem_characterization_non_degeneracy} has a spectacular consequence: non-degeneracy does not depend on the order of optimality (\Cref{corollary_degeneracy_collapse}). 
In more simple terms, be it the identification of gain optimal policies, of bias optimal policies, or of Blackwell optimal policies, consistent algorithms can only stop on the same exact class of Markov decision processes: exactly those that have a unique Bellman optimal policy.

\begin{corollary}[Degeneracy collapse]
	\label{corollary_degeneracy_collapse}
	The set of degenerate models with regards to $\optpolicies_{-1}$, $\optpolicies_0$, $\optpolicies_1$, \dots, $\optpolicies_\infty$ are all the same. 
\end{corollary}

\Cref{theorem_characterization_non_degeneracy} is a strict generalization of Theorem 1 from \cite{boone_identification_2023}, in which the uniqueness of the bias optimal policy is shown necessary for the learnability of Blackwell-optimal policies in deterministic policies. 
Our result extends this to stochastic models, focuses on consistent algorithms, and gives an exact and simpler characterization of the unlearnable models.

The following sections will be dedicated to proving and explaining \Cref{theorem_characterization_non_degeneracy}.

    \section{DEGENERATE MODELS AND STOPPING IN FINITE TIME}
    \label{sect:degen}
    The main result (\Cref{theorem_characterization_non_degeneracy}) claims that for a consistent algorithm to stop in finite time, the Bellman optimal policy of the underlying instance must be unique. 
This section unravels the reason behind this result with \Cref{corollary_non_degenerate_unique_bellman}: if multiple Bellman optimal policies coexist, then a consistent algorithm will indefinitely struggle to determine which one is supposed to be the best. 

\subsection{Non-Degeneracy and Locally Optimal Policies}
\label{section_nondegeneracy}

We begin by drawing a general observation on non-degenerate instances.
Non-degenerate instances admit policies that remain optimal under perturbations of the instance itself. 
This observation does not depend on the optimality order, or even on the optimality notion considered, see \Cref{proposition_non_degenerate_general} below. 

\begin{proposition}
\label{proposition_non_degenerate_general}
    Let $\optpolicies : \models \to 2^\policies$ be a measurable optimality map. 
    If $\model \in \models$ is non-degenerate, then there is a policy $\policy \in \optpolicies(\model)$ that remains optimal in a neighbourhood of $\model$, i.e.,
    \begin{equation*}
        \exists \epsilon > 0,
        \quad
        \bigcap \braces*{
            \optpolicies(\model')
            :
            \mdpdistance(\model, \model') < \epsilon
        }
        \ne
        \emptyset
        .
    \end{equation*}
\end{proposition}
\begin{proof}[Proof Sketch]
    If $\model$ is non-degenerate, there exists a consistent algorithm that identifies an optimal policy $\recommendation$ of $\model$ in finite time. 
    Because the algorithm only takes decisions from observed histories, it is very likely to recommend $\recommendation$ in finite time for every instance $\model'$ that produces histories similar enough to $\model$. 
    In other words, the recommendation of the algorithm is continuous, so that if $\model$ and $\model'$ are close, the algorithm will recommend $\recommendation$ for both $\model$ and $\model'$.
    Accordingly, by consistency, $\recommendation$ is an optimal policy of $\model'$ as well. 
\end{proof}

\begin{figure}[!ht]
	\centering
	\begin{tikzpicture}[scale=.8]
		\filldraw[fill=ForestGreen!20, draw=ForestGreen!20, rounded corners=2] (0,0) -- (5.5,5.5) -- (5.5,0) -- cycle;
		\filldraw[fill=violet!20, draw=violet!20, rounded corners=2] (0,0) -- (5.5,5.5) -- (0,5.5) -- cycle;
		\draw[thick] (0.02,0.02) -- (5.48,5.48);
		\node at (5,5) [above left, rotate = 45] {$\Pi^* = \{\pi_1,\pi_2\}$};
		\node at (1.5,5.5) [below] {{\color{violet}$\Pi^* = \{\pi_1\}$}};
		\node at (4,0) [above] {{\color{ForestGreen}$\Pi^* = \{\pi_2\}$}};
		\fill (2,2) circle (2pt) node[left] {$M$};
		\fill[fill = violet] (1.5,2.5) circle (2pt) node[above left] {{\color{violet}$\widehat M$}};
		\fill[fill = ForestGreen] (2.5,1.5) circle (2pt) node[right] {{\color{ForestGreen}$\widetilde M$}};
	\end{tikzpicture}
	\caption{
        Illustration of \Cref{proposition_non_degenerate_general} --- as long as we can put \textcolor{violet}{$\widehat M$} and \textcolor{ForestGreen}{$\widetilde M$} arbitrarily close to $M$ but having no common optimal policy, we can never learn an optimal policy for $M$
    }
	\label{figure_illustration_halo}
\end{figure}
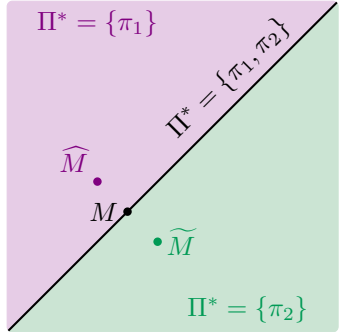

\Cref{proposition_non_degenerate_general} drives the remaining of the proof, and we will work with a converse version of it:
if, for an optimality map $\optpolicies: \models \to 2^\policies$, every $\policy \in \optpolicies(\model)$ can be made into the unique optimal policy of an instance $\model'$ arbitrarily close to $\model$, then $\model$ cannot be non-degenerate unless it has a unique optimal policy. 

\subsection{Making a Bellman Optimal Policy into the Unique Gain Optimal Policy}
\label{section_bellman_to_gain}

Following \Cref{proposition_non_degenerate_general}, to show that non-degenerate instances have a unique Bellman optimal policy, we prove that a Bellman optimal policy can be isolated as the unique gain optimal policy through an ergodification technique, that we call \emph{shattering}. 

\begin{proposition}[Shattering Bellman optimality]
\label{proposition_shattering}
    Let $\model$ be a communicating instance.
    For every unichain Bellman optimal policy $\policy \in \bellmanpolicies(\model)$ and precision $\epsilon > 0$, there exists $\model' \in \models$ with $\mdpdistance(\model, \model') < \epsilon$ for which $\policy$ is the unique gain optimal policy, i.e., $\optpolicies_{-1}(\model') = \braces{\policy}$. 
\end{proposition}

Note that \Cref{proposition_shattering} only claims that \emph{unichain} Bellman optimal policies may be isolated in such fashion --- that detail will be taken care of later on. 

\Cref{proposition_shattering} is derived as a two-step process.
First, we make $\policy \in \bellmanpolicies(\model)$ the \emph{unique} Bellman optimal policy of $\model' \approx \model$ by degrading the mean reward at every state-action pair that $\policy$ does not play (\Cref{lemma_isolating_bellman}).
Then, we argue that if an instance has a unique Bellman optimal policy, then it can be made into the unique gain optimal policy under a well-chosen (infinitesimal) ergodic transform. 
We provide an artist's view of the resulting Markov decision process in \Cref{figure_ergodification}.

\makeatletter
\newcommand*{\shifttext}[2]{%
  \settowidth{\@tempdima}{#2}%
  \makebox[\@tempdima]{\hspace*{#1}#2}%
}
\makeatother

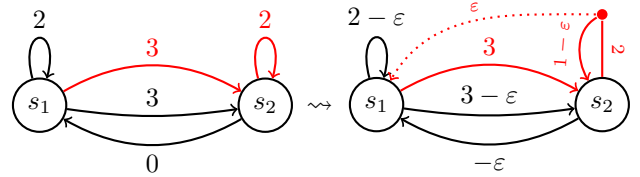
\begin{figure}[ht]
	\begin{tikzpicture}[node distance={30mm},thick,main/.style = {draw,circle}]
		\node[main] (1) {$\state_1$};
		\node[main] (2) [right of=1] {$\state_2$};
		\draw[->] (1) to [loop above,looseness=10] node{$2$} (1);
		\draw[->, color=red] (2) to [loop above,looseness=10] node{$2$} (2);
        \draw[->] (2) to[bend left] node[midway, below] {$0$} (1);
        \draw[->] (1) to[bend right=8] node[midway, above=0pt] {$3$} (2);
        \draw[->, color=red] (1) to[bend left] node[midway, above] {$3$} (2);
	\end{tikzpicture}
    \hfill
    \shifttext{0.3em}{\raisebox{2.6em}{$\leadsto$}}
    \hfill
	\begin{tikzpicture}[node distance={30mm},thick,main/.style = {draw,circle}]
		\node[main] (1) {$\state_1$};
		\node[main] (2) [right of=1] {$\state_2$};
        \node[fill=red, circle, inner sep=0.5mm] (mid) at (30mm, 12mm) {};
		\draw[->] (1) to [loop above,looseness=10] node{$2-\epsilon$} (1);
        \draw[->] (2) to[bend left] node[midway, below] {$-\epsilon$} (1);
        \draw[->] (1) to[bend right=8] node[midway, above=0pt] {$3-\epsilon$} (2);
        \draw[color=red, ->] (1) to[bend left] node[midway, above] {$3$} (2);
        \draw[color=red] (mid) to node[midway, above, sloped] {\scriptsize $2$} (2);
        \draw[color=red, ->] (mid) to[in=90+30,out=180+30] node[midway, above, sloped] {\scriptsize $1-\epsilon$} (2);
        \draw[color=red, ->, dotted] (mid) to[in=60,out=180] node[midway, above, sloped] {\scriptsize $\epsilon$} (1);
	\end{tikzpicture}
    \caption{
    \label{figure_ergodification}
        An example of the shattering operation, making a (non-unique) bias optimal policy (\textcolor{red}{in red}) into the unique gain optimal policy. 
    }
\end{figure}

\begin{lemma}
\label{lemma_isolating_bellman}
    Let $\model \equiv (\pairs, \rewarddistribution, \kernel)$ be a communicating instance.
    For every unichain Bellman optimal policy $\policy \in \bellmanpolicies(\model)$ and $\epsilon > 0$, there exists $\model' \equiv (\pairs, \reward', \kernel)$ with $\mdpdistance(\model, \model') < \epsilon$ for which
    \vspace{-1em} 
    \begin{enum}
        \item $\policy$ is the unique Bellman optimal policy;
        \item $\optgain(\model') = \optgain(\model)$; 
        \item $\optbias(\model') = \optbias(\model)$. 
    \end{enum}
\end{lemma}

\begin{proof}[Proof Sketch of \Cref{lemma_isolating_bellman}]
    We consider the penalized Markov decision process $\model'_\epsilon \equiv (\pairs, \reward'_\epsilon, \kernel)$ with rewards
    \begin{equation*}
        \reward'_\epsilon (\state, \action)
        :=
        \reward_\epsilon (\state, \action)
        - \epsilon
        \indicator{
            \action \ne \policy(\state)
        }
    \end{equation*}
    and check that the family $(\model'_\epsilon)$ works as intended. 
\end{proof}

It is to be noted that the model $\model'$ given by \Cref{lemma_isolating_bellman} is not $[0, 1]$-reward in general. 
In general, this is not an issue because the latter property can be forced \emph{a posteriori} via a linear reward transform. 

In turn, we provide the construction for \Cref{proposition_shattering}.

\begin{proof}[Proof Sketch of \Cref{proposition_shattering}]
    Given a precision $\epsilon > 0$ and a unichain Bellman optimal policy $\policy \in \bellmanpolicies(\model)$, let $\model'$ be the instance isolating $\policy$ as the unique Bellman optimal policy provided by \Cref{lemma_isolating_bellman}. 
    Then, we consider the ergodic transform $\model'' \equiv (\pairs, \reward'', \kernel'')$ with reward and kernel functions given by
    \begin{equation*}
    \begin{aligned}
        \reward'' (\state, \action)
        & := 
        \reward'(\state, \action)
        + \parens*{\kernel''(\state, \action) - \kernel'(\state, \action)} \biasof{\policy}(\model')
        \\
        \kernel''(\state'| \state, \action)
        & :=
        (1 - \epsilon) \kernel'(\state'|\state, \action)
        + \frac{\epsilon}{\abs{\actions(\state)}}
        .
    \end{aligned}
    \end{equation*}
    We check that $\mdpdistance(\model, \model'') \le C \epsilon$ for some $C \ge 0$, and that for $\epsilon$ small enough, we have $\optpolicies_{-1}(\model'') = \braces{\policy}$. 
    To force the rewards to be $[0, 1]$, we further apply the linear reward transform $\varphi(x) := C \epsilon + (1 - 2 C \epsilon) x$.
\end{proof}

\subsection{Non-Degeneracy and Unique Bellman Optimal Policies}

We can almost directly plug \Cref{proposition_non_degenerate_general,proposition_shattering}, but we have one issue to circumvent: \Cref{proposition_shattering} can only shatter \emph{unichain} Bellman optimal policies.
Therefore, we need to prove that if the Bellman optimal policy isn't unique, then there are multiple unichain Bellman optimal policies.
Thankfully, this is indeed true. 

\begin{lemma}
\label{lemma_force_unichain_bellman}
    If a communicating Markov decision process has multiple Bellman optimal policies, then it has multiple unichain Bellman optimal policies. 
\end{lemma}

The proof of \Cref{lemma_force_unichain_bellman} relies on the intimate structure of Bellman optimal policies; we defer it to Appendix. 

\begin{corollary}
\label{corollary_non_degenerate_unique_bellman}
    Given $n \ge -1$, every $\optpolicies_n$-non-degenerate instance has a unique Bellman optimal policy. 
\end{corollary}

\begin{proof}
    Let $\model \in \models$ be a non-degenerate communicating instance.
    Denote $\bellmanpolicies_u (\model)$ its set of unichain Bellman optimal policies.
    By \Cref{proposition_shattering}, for every unichain Bellman optimal policy $\policy \in \bellmanpolicies_u(\model)$ and precision $\epsilon > 0$, there exists $\model_\epsilon^\policy \in \models$ such that $\optpolicies_{-1}(\model_\epsilon^\policy) = \braces{\policy}$ and $\mdpdistance(\model, \model_\epsilon^\policy) < \epsilon$.
    The instance $\model$ is non-degenerate so by \Cref{proposition_non_degenerate_general}, for $\epsilon > 0$ small enough, we have
    \begin{align*}
         \emptyset
         & \ne
         \bigcap \braces*{
             \optpolicies_n (\model')
             :
             \mdpdistance(\model, \model') < \epsilon
         }
         \\
         & \subseteq
         \bigcap_{\policy \in \bellmanpolicies_u (\model)}
         \optpolicies_{-1} (\model^\policy_\epsilon)
         =
         \bigcap_{\policy \in \bellmanpolicies_u (\model)}
         \braces{\policy}
         .
    \end{align*}
    So $\bellmanpolicies_u (\model)$ is necessarily a singleton and thus by \Cref{lemma_force_unichain_bellman}, $\bellmanpolicies (\model)$ is necessarily a singleton as well. 
\end{proof}

    \section{MDPS WITH A UNIQUE BELLMAN OPTIMAL POLICY}
    \label{sect:stopping}
    At this point of the paper, you should be convinced that non-degenerate instances have a unique Bellman optimal policy.
In this section, we show that the converse is equally true, therefore proving \Cref{theorem_characterization_non_degeneracy}.

The heart of the rationale is that if the Bellman optimal policy of $\model$ is unique, then this policy remains the unique optimal policy of instances in a neighbourhood of $\model$. 
Showing that this neighbourhood exists --- and measuring it, is a hard result.
Most of the proof details of the results down below are deferred to Appendix. 

\subsection{Deciding the Uniqueness of Bellman Optimal Policies}

To begin with, we provide an important structural property of Bellman optimal policies that are unique: they have to be unichain (\Cref{lemma_deciding_unique_bellman}). 
This observation is not only very important to the analysis, it also has a huge computational interest, because deciding the uniqueness of the Bellman optimal policy becomes polynomial in time-complexity. 

\begin{lemma}
\label{lemma_deciding_unique_bellman}
    Let $\model$ be a communicating instance. 
    Then

    \vspace{-0.33em} 
    \begin{enum}
        \item
            If $\model$ has a unique Bellman optimal policy, then this policy is unichain.
        \item 
            Conversely, if there exists a unichain policy $\policy \in \policies$ such that $\gainof{\policy}(\state) + \biasof{\policy}(\state) > \reward(\state, \action) + \kernel(\state, \action) \biasof{\policy}$ for all $\action \ne \policy(\state)$ and $\state \in \states$, then $\policy$ is the unique Bellman optimal policy of $\model$. 
    \end{enum}
\end{lemma}

\Cref{lemma_deciding_unique_bellman} provides a simple algorithm to determine whether a communicating Markov decision process has a unique Bellman optimal policy or not. 
First, compute a Bellman optimal policy $\optpolicy$ using Policy Iteration. 
If that policy isn't unichain, reject; otherwise, verify that every gap is positive.
These tests can be realized in time $\OH(\abs{\pairs})$ once $\optpolicy$ is known, which is overall negligble in front of the computation of $\optpolicy$, also polynomial. 

\subsection{A Stopping Rule with \texorpdfstring{$\delta$}{delta}-Probable Correctness}
\label{section_stopping_rule}

Once the Bellman optimal policy is proven to be unichain, we can rely on a range of results that bound the variations of its gain and bias functions with respect to $\mdpdistance(\model, \model')$, that can be found in Appendix.
Without the unichain property, such bounds are impossible to obtain without the assumption that $\model$ and $\model'$ have transition kernels of similar supports, and this assumption is not reasonable in identification problems. 

Now, given a communicating instance $\model \equiv (\pairs, \rewarddistribution, \kernel)$ with unique Bellman optimal policy $\optpolicy$, we introduce the threshold:
\begin{equation}
\label{equation_threshold_non_degenerate}
    \beta(\model)
    :=
    \min \braces*{
        \frac{
            \dmin(\ogaps)
        }{
            (1 + 4 \alpha)
            (2 + \vecspan{\bias^*)}
        }
        ,
        \frac 1{\alpha}
    }
\end{equation}
where we set $\alpha := \min_{\state_\infty \in \states_\infty} \alpha(\state_\infty)$ with $\alpha(\state_\infty) := \max_{\state \in \states} \EE_{\state}^{\optpolicy} \brackets{\inf \braces*{t \ge 1: \State_t = \state_\infty}}$ is the worst hitting time to $\state_\infty$ under the optimal policy $\optpolicy$, and $\states_\infty$ is the set of recurrent states of $\optpolicy$. 
By convention, we set $\beta(\model) := +\infty$ if $\abs{\bellmanpolicies(\model)} > 1$. 

The threshold $\beta(\model)$ provides a simple measure of the radius of the neighbourhood in which all instances have the same Bellman optimal policy as $\model$.
This is formalized in \Cref{lemma_neighborhood_non_degenerate} thereafter.

\begin{lemma}
\label{lemma_neighborhood_non_degenerate}
    Let $\model \equiv (\pairs, \rewarddistribution, \kernel)$ be a communicating instance with unique Bellman optimal policy $\optpolicy$. 
    For every $\model' \equiv (\pairs, \rewarddistribution', \kernel')$ such that $\support(\kernel') \supseteq \support(\kernel)$, if 
    \begin{equation*}
        \mdpdistance(\model, \model')
        <
        \beta(\model)
    \end{equation*}
    then $\optpolicy$ is the unique Bellman optimal policy of $\model'$.
\end{lemma}

From a complexity perspective, the quantity $\beta(\model)$ is computed in time $\OH(\abs{\pairs} + \abs{\states}^4)$. 
Indeed, the recurrent class $\states_\infty$ of $\optpolicy$ is determined in time $\OH(\abs{\states}^2)$.
Given a recurrent state $\state_\infty \in \states_\infty$, the gain and bias functions $\optgain, \optbias$ are obtained by matrix inversions in $\OH(\abs{\states}^3)$ and $\ogaps$ is deduced in $\OH(\abs{\pairs})$. 
Finally, the family of hitting times to $\state_\infty$ (under $\optpolicy$) is the solution of a linear system that can be solved in time $\OH(\abs{\states}^3)$.
Doing it for every $\state_\infty \in \states_\infty$ yields $\alpha$, for a total time cost $\OH(\abs{\states}^4)$.

Since the threshold $\beta(\model)$ is easy to compute, \Cref{lemma_neighborhood_non_degenerate} provides a simple stopping rule for \texttt{HOPE}, our consistent algorithm from \Cref{sect:consistency}. 
First, we design a high probability ball $\braces{\model': \mdpdistance(\hat{\model}_t, \model') < \xi_\delta (t)}$ that contains $\model$ with probability $1 - \delta$ uniformly over time, where the radius $\xi_\delta(t)$ is set to
\begin{equation}
\label{equation_threshold_confidence}
    \xi_\delta 
    (t)
    :=
    \sqrt{
        \frac{
            \abs{\states}
            \log \parens*{
                \frac{2 \abs{\pairs}(1 + t)}\delta
            }
        }{
            \min_{\pair \in \pairs}
            \visits_t (\pair)
        }
    }
    .
\end{equation}
Then, we stop \texttt{HOPE} as soon as $\xi_\delta (t) \le \beta(\hat{\model}_t)$, i.e., when all the instances within the confidence ball share the same unique Bellman optimal policy. 

\begin{proposition}
\label{proposition_hope_stopping_rule}
    Upon running with the stopping rule 
    \begin{equation*}
        \tau_\delta
        :=
        \inf \braces*{
            t \ge 1
            :
            \xi_\delta (t)
            \le 
            \beta (\hat{\model}_t)
            \mathrm{~and~}
            \bellmanpolicies(\hat{\model}_t) = \braces{\recommendation_t}
        }
    \end{equation*}
    where $\xi_\delta(t)$ and $\beta(t)$ are respectively given by \Cref{equation_threshold_confidence,equation_threshold_non_degenerate}, \texttt{HOPE} is $\delta$-PC for $\optpolicies_n$, whatever $n \ge 1$.
    Moreover, for every instance $\model$ with unique Bellman optimal policy, we have $\Pr^{\model, \texttt{HOPE}} (\tau_\delta < \infty) = 1$. 
\end{proposition}

With this, we have all the ingredients necessary to finish our proof.

\subsection{Instances with Unique Bellman Optimal Policies are Non-Degenerate}

At last, the non-degeneracy of instances with unique Bellman optimal policies is a direct consequence of \Cref{proposition_hope_stopping_rule}, concluding the proof of \Cref{theorem_characterization_non_degeneracy}. 

\begin{corollary}
    If a communicating Markov decision process has a unique Bellman optimal policy, then it is $\optpolicies_{n}$-non-degenerate for all $n \ge -1$. 
\end{corollary}

\begin{proof}
    By \Cref{th_consistency}, \texttt{HOPE} is consistent for $\optpolicies_n$. 
    By \Cref{proposition_hope_stopping_rule}, once enriched with the stopping rule $\tau_\delta$ of \Cref{proposition_hope_stopping_rule}, it is $\delta$-PC.
    Lastly, by \Cref{proposition_hope_stopping_rule} again, we have $\Pr^{\model, \texttt{HOPE}}(\tau_\delta < \infty) = 1$ for every instance $\model$ with unique Bellman optimal policy. 
\end{proof}

    \section{CONCLUSION}

As the main take-away from this paper, Bellman optimality is a fundamental limit to identifying $n$-optimal policies. This collapsing result bodes badly for learning in average reward: can we never try to learn policies optimal in both the long and the short term, then? There is hope in the PAC setting, in which one investigates \textit{near} optimal policies.

Another obvious research path, now that we have isolated the models in which optimality is learnable, would be to investigate the necessary and achievable sample complexities.

    \section*{ACKNOWLEDGMENTS}
    The authors acknowledge the funding of the French National Research Agency under the project FATE (ANR22-CE23-0016-01), the PEPR IA FOUNDRY project (ANR-23-PEIA-0003), the ANR LabEx CIMI (grant ANR-11-LABX-0040) within the French State Programme Investissements d'Avenir, and the AI Interdisciplinary Institute ANITI, funded by the France 2030 program under the Grant agreement n°ANR-23-IACL-0002.
    The second author is member of the Inria team Scool.

    \bibliography{bibliography.bib}

    \clearpage
    \appendix
    \onecolumn
    \numberwithin{theorem}{section}
    \numberwithin{definition}{section}
    \numberwithin{equation}{section}

    \section{PRELIMINARY TECHNICAL RESULTS}
    
In this Appendix, we provide a variety of preliminary results.

We start with some concentration results in \Cref{subsection_concentration}, quantifying the speed at which the empirical MDP $\hat{M}_t$ converges towards the real MDP in terms of the distance $\mdpdistance$.
We then give in \Cref{subsection_numerical_res} a numerical result that we'll use in the following.
In \Cref{subsection_visitcounts}, we study the number of visits in each state-action pair under uniform exploration. Finally, for two MDPs with ``small" distance, we study in \Cref{subsection_distances} how closely their bias vectors are, for unichain policies and in general.

\subsection{A Few Concentration Results from Probability Theory}\label{subsection_concentration}

\begin{lemma}[Maximal Azuma]\label{lem_max_azuma}
	Let $(X_m)$ be a martingale difference sequence, with almost surely $X_m$ bounded between $-0.5$ and $0.5$ for all $m\in \NN$.
    Denote $\hat{\mu}_n:= \frac{1}{n}\sum_{m=1}^n X_m$ the empirical mean of the first $n$ variables. 
    For all $x\geq 0$, we have 
    \begin{equation*}
        \Pr \parens*{
            \exists m \geq n, 
            \hat{\mu}_m\geq x
        }
        \leq 
        \exp(-2nx^2)
        .
    \end{equation*}
\end{lemma}

\begin{lemma}[Time-uniform Weissman, \cite{boone_thesis_2024}]\label{lem:devrew}
	Let $q$ be a distribution over $\{1,\dots,d\}$. 
    Let $(X_m)$ be a sequence of i.i.d.~random variables of distribution $q$, and $\hat{q}_n$ be the empirical mean over $n$ samples, $\frac{1}{n} \sum_{m=1}^n X_m$. 
    Then, for all $\delta >0$,
    \begin{equation*}
        \Pr \parens*{
            \exists n\geq 1,
            \norm{\hat{q}_n -q}_1 
            \geq 
            \sqrt{ 
                \frac{d\log \parens*{(2\sqrt{1+n})/\delta}}{n}
            }
        }
        \leq 
        \delta
        .
    \end{equation*}
\end{lemma}

\begin{lemma}[Time-uniform Azuma, Lemma 5 of \cite{bourel_tightening_2020}]\label{lem:devtrans}
	Let $(Y_m)$ be a sequence of i.i.d.~random variables bounded between 0 and 1, with $\hat{\mu}_n$ the empirical mean estimate after $n$ samples and $\mu$ the real mean.
    Then, for all $\delta>0$, 
    \begin{equation*}
        \Pr \parens*{
            \exists n \ge 1,
            \abs*{\hat{\mu}_n-\mu}
            \geq
            \sqrt{\left( 1+\frac{1}{n}\right) \frac{\log\parens*{\sqrt{1+n}/\delta}}{2n}}
            \leq
            \delta 
        }
    \end{equation*}
\end{lemma}
 
\begin{corollary}[Time-uniform MDP concentration]\label{cor:convergencespeed}
    Let $\model$ be a Markov decision process with pair and state spaces $\pairs$ and $\states$ respectively. 
    Let $\hat{\model}_t$ be its empirical estimation, as given in (...).
    For all $\delta > 0$, we have
	\begin{equation*}
        \Pr \parens*{
            \exists t \ge 1,
            \mdpdistance(\model, \hat{\model}_t)
            \geq
            \sqrt{
                \frac{
                    |\states|
                    \log \parens*{
                        \parens{4|\pairs|\sqrt{1+\min \visits_t}}
                        /
                        {\delta}
                    }
                }{\min \visits_t}
            }
        } 
        \le
        \delta
    \end{equation*}
\end{corollary}

\begin{proof}
    Write $\model \equiv (\pairs, \kernel, \reward)$. 
    Let $\xi(n, \delta) := \sqrt{\abs{\states} \log((4 \abs{\pairs} \sqrt{1 + n})/\delta) / n}$ the error as given by \Cref{cor:convergencespeed}. 
    We have
	\begin{align*} 
        \Pr \parens*{
            \exists t \in \NN, 
            \mdpdistance(\model, \hat{\model}_t)
            \geq
            \xi(\min(\visits_t), \delta) 
        }
        & \overset{\eqnum{1}}\leq 
        \sum_{(s,a) \in \pairs} 
        \brackets*{
            \begin{subarray}{c}
                \displaystyle
                \Pr \parens*{
                    \exists t\in\NN, 
                    \norm*{\hat{\kernel}_t (\state, \action) - \kernel(\state, \action)}_1
                    \geq
                    \xi(\min(\visits_t), \delta)
                }
                \\
                \displaystyle
                +
                \\
                \displaystyle
                \Pr \parens*{
                    \exists t\in\NN, 
                    \abs*{\hat{\reward}_t (\state, \action) - \reward(\state, \action)}
                    \geq
                    \xi(\min(\visits_t), \delta)
                }
            \end{subarray}
        }
        \\
        & \overset{\eqnum{2}}\leq 
        \sum_{(s,a) \in \pairs} 
        \brackets*{
            \begin{subarray}{c}
                \displaystyle
                \Pr \parens*{
                    \exists t\in\NN, 
                    \norm*{\hat{\kernel}_t (\state, \action) - \kernel(\state, \action)}_1
                    \geq
                    \xi(\visits_t (\state, \action), \delta)
                }
                \\
                \displaystyle
                +
                \\
                \displaystyle
                \Pr \parens*{
                    \exists t\in\NN, 
                    \abs*{\hat{\reward}_t (\state, \action) - \reward(\state, \action)}
                    \geq
                    \xi(\visits_t (\state, \action), \delta)
                }
            \end{subarray}
        }
        \\
        & \overset{\eqnum{3}}\le
        \abs{\pairs} \cdot \frac \delta{2\abs{\pairs}}
        +
        \abs{\pairs} \cdot \frac \delta{2\abs{\pairs}}
        = \delta
	\end{align*}
    where 
    \texteqnum{1} unfolds the definition of $\mdpdistance(-)$ and uses a union-bound over the set of pairs;
    \texteqnum{2} uses that, for $x>0$ and $y>1$, $x\mapsto \frac 1x \cdot \log y\sqrt{1+x}$ is decreasing, so that $\min \visits_t$ can be changed to $\visits_t (\state, \action)$;
    and
    \texteqnum{3} bounds the deviations of transition kernels with \Cref{lem:devrew} and the deviations of rewards with \Cref{lem:devtrans}).
\end{proof}

\subsection{A Numerical Result}\label{subsection_numerical_res}

We will also be using the following folklore bound. 

\begin{lemma}[Folklore]
	\label{lemma_l1_deviation_span}
	Let $p, q$ be two probability distributions over $\braces{1, ..., d}$ and let $u \in \RR^d$.
	Then $\abs{(q - p) u} \le \frac 12 \vecspan{u} \norm{q - p}_1$.
\end{lemma}
\begin{proof}
	Let $\unit := (1, \ldots, 1)$ be the unitary vector.
	Because $\sum_{i=1}^d p_i = \sum_{i=1} q_i$, we have $(q - p) u = (q - p) (u + \lambda e)$ for all $\lambda \in \RR$. 
	We find
	\begin{equation*}
		\abs{(q - p) u}
		=
		\inf_{\lambda \in \RR}
		\abs*{
			(q - p) (u + \lambda e)
		}
		\overset{(\dagger)}\le
		\inf_{\lambda \in \RR}
		\norm{q - p}_1 
		\norm{u + \lambda e}_\infty
		\overset{(\ddagger)}=
		\frac 12 \vecspan{u}
		\norm{q - p}_1 
	\end{equation*}
	where 
	$(\dagger)$ follows from H\"older's inequality; and
	$(\ddagger)$ follows by setting $\lambda := - \frac 12 (\min(u) + \max(u))$. 
\end{proof}

\subsection{Visit Counts under Uniform Exploration}\label{subsection_visitcounts}

Algorithm \texttt{HOPE} uses uniform exploration. We thus need to determine how often each pair is visited, to use in conjunction with \Cref{cor:convergencespeed}.

\begin{lemma}\label{lem_visits}
	Let $M\equiv (\pairs,p,r)$ a communicating Markov decision process and let $\pi$ be the uniformly random policy, given by $\pi(a|s) := |\actions(s)|^{-1}$. There exist constants $\beta,\gamma,\lambda>0$ such that \[\Pr^{M,\pi}(\exists t\geq T,\exists z\in \pairs, \visits_t(z) < \beta t)\leq \exp(-\gamma T+\lambda)\]
\end{lemma}

\begin{proof}
	Fix a pair $\pair_0\in\pairs$. Consider the deterministic reward function $f(\pair):=\ind(\pair=\pair_0)$, and let $M_{\pair_0}:=(\pairs,p,f)$ the copy of $M$ with reward function $f$. In $M_{\pair_0}$, we get a reward only when we sample $\pair_0$. Let $\gain_{\pair_0}^\pi$, $\bias_{\pair_0}^\pi$ and $\gaps_{\pair_0}^\pi$ the gain, 0-bias and 0-gap functions of $\pi$ in $M_{\pair_0}$. Because $M_{\pair_0}$ is communicating and $\pi$ is uniform, every pair $\pair$ is recurrent under $\pi$, so in particular $\pair_0$ is recurrent. Therefore, $\gain_{\pair_0}(\pair)>0$. Set $\gamma_{\pair_0} :=\max\{ \vecspan{\bias_{\pair_0}},\norm{\gaps^\star_{\pair_0}}_\infty \}$. We have \begin{align*}
		\visits_{t}^{M} (\pair_0) &= N_{t}^{M_{\pair_0}} (\pair_0) = \sum_{i=1}^t(\Pair_i =\pair_0)\\
		& \overset{\eqnum{1}}{=} \sum_{i=1}^t \left( \gain_{\pair_0} + \bias_{\pair_0}(\State_i) -p(\Pair_i) \bias_{\pair_0} - \gaps_{\pair_0}(\Pair_i)\right) \\
		& \overset{\eqnum{2}}{\geq} t\gain_{\pair_0} +\bias_{\pair_0}(\State_1) - p(Z_t)\bias_{\pair_0} +\sum_{i=1}^t \left( e_{\State_{i+1}} - p(\Pair_i)\right) \bias_{\pair_0} - \sum_{i=1}^t \left( e_{\Action_i} -\pi(\State_{i-1})\right) \gaps_{\pair_0}(\State_i) \\
		& \geq t\gain_{\pair_0} -\gamma_{\pair_0} +\sum_{i=1}^t\left( (e_{\State_{i+1}} -p(\Pair_i))\bias_{\pair_0} + (e_{\Action_i} -\pi(\State_{i-1})) \gaps_{\pair_0} (\State_i) \right)
	\end{align*} where $\texteqnum{1}$ comes from the definition of $\Delta_{\pair_0}$ ; $\texteqnum{2}$ uses that $\pi(s) \cdot\Delta_{\pair_0}(s)=0$ for any state $s$, where $\gaps_{\pair_0}(s) := \left(\gaps_{\pair_0}(s,a)\right)_{a\in\actions_s}$. In the last inequation, we recognize in the right hand side term a sum of martingale differences, each bounded by $2\gamma_{\pair_0}$. Therefore, by \Cref{lem_max_azuma} applied to \[(\nu_i)_i:= \left( \frac{(e_{\State_{i+1}} -p(\Pair_i))\bias_{\pair_0} + (e_{\Action_i} -\pi(\State_{i-1})) \gaps_{\pair_0} (\State_i)}{4\gamma_{\pair_0}} \right)_i\] we get for any $x\geq 0$ \begin{align*}
	\exp\left( -\frac{Tx^2}{8\gamma_{\pair_0}^2}\right)&\geq \Pr \left( \exists t\geq T, \frac{1}{t} \sum_{i=1}^t \nu_i \leq - \frac{x}{4\gamma_{\pair_0}}\right) \\
	&\geq \Pr\left( \exists t\geq T, \sum_{i=1}^t (e_{\State_{i+1}} -p(\Pair_i))\bias_{\pair_0} + (e_{\Action_i} -\pi(\State_{i-1})) \gaps_{\pair_0} (\State_i) \leq -xt\right) \\
	&\geq \Pr\left(\exists t\geq T, N_{t}(\pair_0) \leq t\gain_{\pair_0}-\gamma_{\pair_0}-xt \right)
	\end{align*}
	and we get the result by picking $x := \frac{1}{3}\min_\pair g_\pair >0$, with $\beta := x$, $\gamma := \frac{(\min_\pair \gain_\pair)^2}{72(\max_\pair \gamma_\pair)^2}$ and $\lambda:= \frac{3\gamma \max_\pair \gamma_\pair}{\min_\pair \gain_\pair} + \ln |\pairs|$. 
\end{proof}

\subsection{Distances Between Two MDPs}\label{subsection_distances}

For two MDPs, it is expected that, as they get close, their behaviour should also become similar: gap functions, bias functions... In the following, we describe the variation of those quantities with regards to the distance between the models. We start with the special case of unichain policies.

\subsubsection{Unichain Policies}

The goal of this paragraph is to provide a bound on the variations of the gap function $\gapsof{\policy}(\state, \action) := \gainof{\policy}(\state) + \biasof{\policy}(\state) - \kernel(\state, \action) \biasof{\policy} - \reward(\state, \action)$ of a \emph{unichain} policy $\policy$ under changes of the reward $\reward$ and kernel function $\kernel$ (\Cref{lemma_unichain_gap_variations}).
Recall that the distance between model $\model$ and $\model'$ in $\ell_\infty$-norm is given by
\begin{equation*}
	\distance_\infty (\model, \model')
	:=
	\max_{\pair \in \pairs} \braces*{
		\abs{\reward'(\pair) - \reward(\pair)}
		+ 
		\norm{\kernel'(\pair) - \kernel(\pair)}_1
	}
	.
\end{equation*}
Because the gap function is a combination of the gain and the bias function, the result is obtained by combining deviational bounds on these to quantities (\Cref{lemma_unichain_gain_variations} and \Cref{lemma_unichain_bias_variations}). 
The technique is adapted and generalized from \cite[Appendix~D]{boone_asymptotically_2025}.

\begin{lemma}[Variations of the gain, unichain case, \cite{boone_logarithmic_2025}]
	\label{lemma_unichain_gain_variations}
	Let $\model \equiv (\pairs, \kernel, \reward)$ be a Markov decision process and fix $\policy \in \policies$ a unichain policy of $\model$.
	For all $\model' \equiv (\pairs, \kernel', \reward')$, we have
	\begin{equation*}
		\norm*{\gainof{\policy}(\model') - \gainof{\policy}(\model)}_\infty
		\le
		(1 + \vecspan{\biasof{\policy}(\model)} 
		\distance_\infty (\model, \model')
		.
	\end{equation*}
\end{lemma}

\begin{lemma}[Variations of reaching times, unichain case]
	\label{lemma_unichain_reaching_time_variations}
	Let $\model \equiv (\pairs, \kernel, \reward)$ be a Markov decision process and fix $\policy \in \policies$ a unichain policy of $\model$.
	Let $\state_\infty \in \states$ be a recurrent state of $\policy$ and denote $\tau_{\infty} := \inf \braces{t \ge 1 : \State_t = \state_\infty}$ the reaching time to $\state_\infty$. 
	For every model $\model' \equiv (\pairs, \kernel', \reward')$ such that $\support(\kernel') \supseteq \support(\kernel)$, for all $\state \in \states$, we have
	\begin{equation*}
		\abs*{
			\EE_{\state}^{\model', \policy} [\tau_\infty]
			- \EE_{\state}^{\model, \policy} [\tau_\infty]
		}
		\le
		\frac 12
		\EE_{\state}^{\model', \policy} [\tau_\infty]
		\max_{\state' \in \states} 
		\braces*{
			\EE_{\state'}^{\model, \policy} [\tau_\infty]
		}
		\distance_\infty (\model, \model')
		.
	\end{equation*}
	In particular, if $\distance_\infty (\model, \model') \le \parens*{\max_{\state' \in \states} \EE_{\state'}^{\model, \policy}[\tau_\infty]}^{-1}$, then $\EE_{\state}^{\model', \policy} [\tau_\infty] \le 2 \EE_{\state}^{\model, \policy}[\tau_\infty]$ for all $\state \in \states$.
\end{lemma}

\begin{proof}
	Let $\model_0 \equiv (\pairs, \kernel_0, \reward_0)$ be the model with reward function $\reward_0 (\state, \action) := \indicator{\state \ne \state_\infty}$ and kernel
	\begin{equation*}
		\kernel_0 (\state'|\state, \action)
		:= 
		\begin{cases}
			\kernel (\state'|\state, \action) & \text{if $\state \ne \state_\infty$;} \\
			0 & \text{if $\state = \state_\infty$ and $\state' \ne \state_\infty$;} \\
			1 & \text{if $\state \ne \state_\infty$ and $\state' = \state_\infty$.} \\
		\end{cases}
	\end{equation*}
	Define $\model'_0$ similarly. 
	We see that $\state_\infty$ is eventually absorbing under $\policy$ in $\model_0$; This is also true in $\model'_0$ since $\support(\kernel') \supseteq \support(\kernel)$. 
	So $\gainof{\policy}(\model_0) = \gainof{\policy}(\model'_0) = 0$ and $\biasof{\policy}(\state_\infty; \model_0) = \biasof{\policy}(\state_\infty; \model'_0) = 0$.
	Now, for $\state \in \states$, we have
	\begin{align*}
		\EE_{\state}^{\model', \policy}[\tau_\infty]
		& =
		\EE_{\state}^{\model'_0, \policy} \brackets*{
			\sum_{t=1}^{\tau_\infty-1}
			\reward_0 (\state \ne \state_\infty)
		}
		\\
		& \overset{\eqnum{1}}=
		\EE_{\state}^{\model'_0, \policy} \brackets*{
			\sum_{t=1}^{\tau_\infty-1}
			\parens*{
				\gainof{\policy}(\model_0)
				+ \parens*{
					\unit_{\State_t}
					- \kernel(\Pair_t)
				} \biasof{\policy}(\model_0)
			}
		}
		\\
		& =
		\biasof{\policy}(\state; \model_0) - \biasof{\policy}(\state_\infty; \model_0)
		+
		\EE_{\state}^{\model'_0, \policy} \brackets*{
			\sum_{t=1}^{\tau_\infty-1}
			\parens*{
				\kernel' (\Pair_t)
				- \kernel(\Pair_t)
			} \biasof{\policy}(\model_0)
		}
	\end{align*}
	where $\texteqnum{1}$ follows from the Poisson equation of $\policy$ in $\model_0$.
	Similarly, by unfolding the Poisson equation of $\policy$ in $\model''_0$, we find that $\EE_{\state}^{\model'', \policy}[\tau_\infty] = \biasof{\policy}(\state; \model''_0)$ for every $\model'' \in \braces{\model, \model'}$. 
	Applying \Cref{lemma_l1_deviation_span}, we conclude that
	\begin{align*}
		\abs*{
			\EE_{\state}^{\model', \policy} [\tau_\infty]
			- \EE_{\state}^{\model, \policy} [\tau_\infty]
		}
		& \le 
		\frac 12 \vecspan{\biasof{\policy}(\model_0)}
		\EE_{\state}^{\model', \policy} [\tau_\infty]
		\norm{\kernel' - \kernel}_\infty
		\\
		& \le
		\frac 12
		\EE_{\state}^{\model', \policy} [\tau_\infty]
		\max_{\state' \in \states} 
		\braces*{
			\EE_{\state'}^{\model, \policy} [\tau_\infty]
		}
		\distance_\infty (\model, \model')
		.
	\end{align*}
	This concludes the proof.
\end{proof}

\begin{lemma}[Variations of the bias, unichain case]
	\label{lemma_unichain_bias_variations}
	Let $\model \equiv (\pairs, \kernel, \reward)$ be a Markov decision process and fix $\policy \in \policies$ a unichain policy of $\model$.
	Let $\state_\infty \in \states$ be a recurrent state of $\policy$ and denote $\tau_{\infty} := \inf \braces{t \ge 1 : \State_t = \state_\infty}$ the reaching time to $\state_\infty$. 
	For every model $\model' \equiv (\pairs, \kernel', \reward')$ such that $\support(\kernel') \supseteq \support(\kernel)$ and for all $\state \in \states$, we have
	\begin{equation*}
		\abs*{
			\biasof{\policy}(\state; \model') - \biasof{\policy}(\state; \model)
			+ \biasof{\policy}(\state_\infty; \model) - \biasof{\policy}(\state_\infty; \model')
		}
		\le
		4 \EE_{\state}^{\model, \policy} [\tau_\infty] \parens*{
			1 + \vecspan{\biasof{\policy}(\model)}
		} \distance_\infty (\model, \model')
	\end{equation*}
	provided that $\distance_\infty (\model, \model') \le \parens*{\max_{\state' \in \states} \EE_{\state'}^{\model, \policy}[\tau_\infty]}^{-1}$.
\end{lemma}

\begin{proof}
	Note that since $\support(\kernel') \supseteq \support(\kernel)$ and that $\policy$ is unichain in $\model$, it is unichain in $\model'$ as well.
	We further have $\EE_{\state}^{\model', \policy}[\tau_\infty] < \infty$ regardless of the initial state $\state \in \states$.
	Now, fix $\state \in \states$. 
	We have
	\begin{align*}
		& \biasof{\policy}(\state; \model')
		- \biasof{\policy}(\state_\infty; \model')
		\\
		& :=
		\EE_{\state}^{\model', \policy} \brackets*{
			\sum_{t=0}^{\tau_\infty-1}
			\parens*{
				\reward'(\Pair_t) - \gainof{\policy}(\model')
			}
		}
		\\
		& = 
		\EE_{\state}^{\model', \policy} \brackets*{
			\sum_{t=0}^{\tau_\infty-1}
			\parens*{
				\reward(\Pair_t) - \gainof{\policy}(\model)
			}
		}
		+ 
		\EE_{\state}^{\model', \policy} \brackets*{
			\sum_{t=0}^{\tau_\infty-1}
			\parens*{
				\reward'(\Pair_t) - \reward(\Pair_t)
			}
		}
		+
		\EE_{\state}^{\model', \policy} \brackets*{
			\sum_{t=0}^{\tau_\infty-1}
			\parens*{
				\gainof{\policy}(\model) - \gainof{\policy}(\model')
			}
		}
		\\
		& \overset{\eqnum{1}}=
		\EE_{\state}^{\model', \policy} \brackets*{
			\sum_{t=0}^{\tau_\infty-1}
			\parens*{
				\kernel(\Pair_t) \biasof{\policy}(\model)
				- \biasof{\policy}(\State_t; \model)
			}
		}
		+ \dd \reward
		+ \dd \gain
		\\
		& \overset{\eqnum{2}}=
		\biasof{\policy}(\state; \model) - \biasof{\policy}(\state_\infty; \model)
		+
		\EE_{\state}^{\model', \policy} \brackets*{
			\sum_{t=0}^{\tau_\infty-1}
			\parens*{
				\kernel(\Pair_t) - \kernel'(\Pair_t) 
			} \biasof{\policy}(\model)
		}
		+ \dd \reward
		+ \dd \gain
		\\
		& \overset{\eqnum{3}}\equiv
		\biasof{\policy}(\state; \model) - \biasof{\policy}(\state_\infty; \model)
		+ \dd \reward
		+ \dd \gain
		+ \dd \kernel
	\end{align*}
	where 
	$\texteqnum{1}$ follows from the Poisson equation of $\policy$ in $\model$ and introduces $\dd \reward$ and $\dd \gain$ for the two noise terms;
	$\texteqnum{2}$ is obtained by recognizing a telescopic sum; and
	$\texteqnum{3}$ introduces $\dd \kernel$ as the last noise term. 
	
	To bound each term, notice that since $\distance_\infty (\model, \model') \le 1 / \max_{\state' \in \states} \EE_{\state'}^{\model, \policy} [\tau_\infty]$, we have $\EE_{\state}^{\model', \policy}[\tau_\infty] \le 2 \EE_{\state}^{\model, \policy}[\tau_\infty]$ (\Cref{lemma_unichain_reaching_time_variations}).
	Combined with the bound on gain deviations of \Cref{lemma_unichain_gain_variations} and \Cref{lemma_l1_deviation_span}, we obtain the desired result.
\end{proof}

\begin{lemma}[Variations of the gap function, unichain case]
	\label{lemma_unichain_gap_variations}
	Let $\model \equiv (\pairs, \kernel, \reward)$ be a Markov decision process and fix $\policy \in \policies$ a unichain policy of $\model$.
	Let $\state_\infty \in \states$ be a recurrent state of $\policy$ and denote $\tau_{\infty} := \inf \braces{t \ge 1 : \State_t = \state_\infty}$ the reaching time to $\state_\infty$. 
	For every model $\model' \equiv (\pairs, \kernel', \reward')$ such that $\support(\kernel') \supseteq \support(\kernel)$, we have
	\begin{equation*}
		\norm*{
			\gapsof{\policy}(\model')
			- \gapsof{\policy}(\model')
		}_\infty
		\le 
		\parens*{
			1 + 4 \max_{\state' \in \states} \EE_{\state'}^{\model, \policy} [\tau_\infty]
		}
		\parens*{
			2 + \vecspan{\biasof{\policy}(\model)}
		} \distance_\infty (\model, \model')
		.
	\end{equation*}
	provided that $\distance_\infty (\model, \model') \le \parens*{\max_{\state' \in \states} \EE_{\state'}^{\model, \policy}[\tau_\infty]}^{-1}$.
\end{lemma}

\begin{proof}
	Fix $(\state, \action) \in \pairs$. 
	We have
	\begin{align*}
		& \abs*{
			\gapsof{\policy}(\state, \action; \model')
			- \gapsof{\policy}(\state, \action; \model)
		}
		\\
		& =
		\abs*{
			\gainof{\policy}(\state; \model')
			- \gainof{\policy}(\state; \model)
		}
		+
		\abs*{
			\reward'(\state, \action) - \reward(\state, \action)
		}
		+ 
		\abs*{
			\parens*{
				\unit_\state - \kernel'(\state, \action)
			} \biasof{\policy}(\model')
			-
			\parens*{
				\unit_\state - \kernel(\state, \action)
			} \biasof{\policy}(\model)
		}
		\\
		& \overset{\eqnum{1}}\le
		\parens*{
			2 + \frac 12 \vecspan{\biasof{\policy}(\model)}
		} \distance_\infty (\model, \model')
		+ 
		\abs*{
			\parens*{
				\unit_\state - \kernel'(\state, \action)
			} 
			\parens*{
				\biasof{\policy}(\model')
				- \biasof{\policy}(\model)
			}
		}
		+
		\abs*{
			\parens*{
				\kernel'(\state, \action) - \kernel(\state, \action)
			} \biasof{\policy}(\model)
		}
		\\
		& \overset{\eqnum{2}}\le
		\parens*{
			2 + \vecspan{\biasof{\policy}(\model)}
		} \distance_\infty (\model, \model')
		+ 
		\inf_{\lambda \in \RR}
		\abs*{
			\parens*{
				\unit_\state - \kernel'(\state, \action)
			} 
			\parens*{
				\biasof{\policy}(\model')
				- \biasof{\policy}(\model)
				+ \lambda \unit
			}
		}
		\\
		& \overset{\eqnum{3}}\le
		\parens*{
			1 + 4 \max_{\state' \in \states} \EE_{\state'}^{\model, \policy} [\tau_\infty]
		}
		\parens*{
			2 + \vecspan{\biasof{\policy}(\model)}
		} \distance_\infty (\model, \model')
	\end{align*}
	where
	$\texteqnum{1}$ follows by \Cref{lemma_unichain_gain_variations};
	$\texteqnum{2}$ uses \Cref{lemma_l1_deviation_span} to bound the left-most term; and
	$\texteqnum{3}$ chooses $\lambda = \biasof{\policy}(\state_\infty; \model') - \biasof{\policy}(\state_\infty; \model)$ and invokes \Cref{lemma_unichain_bias_variations}.
\end{proof}

\subsubsection{General Case}

For the general case, we import a result from \cite{boone_asymptotically_2025}, see \Cref{lemma_variations_bias} below, that bounds the variations of the bias function between two chains in the general setting. 
In the general setting, this bound requires that $\support(\kernel_1) = \support(\kernel_2)$, i.e., that the transition kernel of the two models have the same support.
It further requires that $\norm{\kernel_2 - \kernel_1}$ is small relatively to the \strong{generalized diameter} of $\kernel_1$, that we recall below.

\begin{definition}[Generalized diameter of a kernel]
\label{definition_kernel_diameter}
    Let $\kernel$ be a transition kernel over some finite space $\states$.
    Let $\states_1, \ldots, \states_m \subseteq \states$ be the recurrent components of the Markov chain induced by $\kernel$ over $\states$.
    The \strong{generalized diameter of $\kernel$} is 
    \begin{equation}
        \diameter(\kernel)
        :=
        \max_{\state_0 \in \states}
        \max_{(\state_i) \in \product_i (\states_i)}
        \EE_{\state}^{\kernel} \brackets*{
            \inf \braces*{
                t \ge 1:
                \State_t \in \braces*{\state_1, \ldots, \state_m}
            }
        }
        .
    \end{equation}
\end{definition}

That is, the generalized diameter is the worst hitting time to a covering of the recurrent components.
Note that $\diameter(\kernel) < \infty$ as soon as $\states$ is finite.
Because, given a MDP, a policy $\policy \in \policies$ induces a kernel $\kernel_\policy$, we will write $\diameter(\policy) \equiv \diameter(\kernel_\policy)$ in abuse of notation.
We finally define the \strong{worst diameter} of a MDP as the largest generalized diameter over its policies, see below.

\begin{definition}
\label{definition_worst_diameter}
    Let $\model$ be a Markov decision process with finitely many states.
    Its \strong{worst diameter} is 
    \begin{equation}
        \worstdiameter(\model) 
        :=
        \max_{\policy \in \policies}
        \diameter(\policy).
    \end{equation}
\end{definition}

The worst diameter is always finite. 
Then, we prove the following.

\begin{lemma}\label{lemma_close_MDPs}
	Let $M \equiv (\pairs,p,r)$ a communicating Markov decision process. For all $n\geq 1$, there exists a constant $\alpha_n(M) \in \RR$ such that, for all $\hat{M}$ with $\mdpdistance^*(\model, \hat{\model}) \le 1/\worstdiameter(\model)$, we have
	\begin{equation}
		\label{equation_close_MDPs}
		\max_{\policy \in \policies} 
		\max_{\pair \in \pairs} 
		\abs*{
			r_n(\pair) 
			+ p(\pair)\nbias_n^\pi 
			- \hat{r}_n(\pair) 
			-\hat{p}(\pair) \hat{\nbias}_n^\pi
		}
		\leq 
		\alpha_n(M) \mdpdistance^* (\model, \hat{\model})
	\end{equation}
\end{lemma}

\begin{lemma}[{Variations of the bias function of a MRP, \cite[Lemma~D.14]{boone_asymptotically_2025}}]
	\label{lemma_variations_bias}
	Let $(\model_i)_{i=1,2} \equiv (\kernel_i, \reward_i)_{i = 1, 2}$ be a pair of MRPs with common state space $\states$. 
	Denote $\bias_i \in \RR^\states$ the bias function of $(\kernel_i, \reward_i)$. 
	If $\norm{\kernel_2 - \kernel_1}_\infty^* \le 1/\diameter(\kernel_1)$, then
	\begin{equation*}
		\norm{\bias_2 - \bias_1}_\infty
		\le 
		\parens*{
			12 
			+ \parens*{
				16 + \abs{\states}
			} \diameter (\kernel_1)
		} \diameter(\kernel_1)
		\mdpdistance(\model_1, \model_2)
	\end{equation*}
	where $\mdpdistance(\model_1, \model_2) := \max \braces*{\norm{\reward_2 - \reward_1}_\infty, \norm{\kernel_2 - \kernel_1}_\infty}$. 
\end{lemma}

\begin{proof}[Proof of \Cref{lemma_close_MDPs}]
	Because $\policies$ is a finite set, it is enough to establish \Cref{equation_close_MDPs} for a single policy $\policy \in \policies$. 
	To begin with, note that by triangular inequality, we have
	\begin{align*}
		\max_{\pair \in \pairs}
		\abs*{
			\reward_n (\pair) 
			+ \kernel(\pair) \nbias_n^\policy
			- \hat{\reward}_n (\pair)
			- \hat{\kernel}(\pair) \hat{\nbias}_n^\policy
		}
		& \le
		\norm*{\reward_n - \hat{\reward}_n}_\infty
		+
		\max_{\pair \in \pairs}
		\abs*{
			(\kernel(\pair) - \hat{\kernel}(\pair))
			\nbias_n^\policy
		}
		+ 
		\max_{\pair \in \pairs}
		\abs*{
			\hat{\kernel}(\pair)
			\parens*{\nbias_n^\policy - \hat{\nbias}_n^\policy}
		}
		\\
		& \le
		\parens*{
			1 + \frac 12 \vecspan{\nbias_n^\policy}
		} \mdpdistance(\model, \hat{\model})
		+
		\norm*{
			\nbias_n^\policy - \hat{\nbias}_n^\policy
		}_\infty
	\end{align*}
	so that the result is about proving that there exists $\alpha'_n (\model) < \infty$ such that $\norm{\nbias_n^\policy - \hat{\nbias}_n^\policy} \le \alpha'_n (\model) \mdpdistance(\model, \hat{\model})$. 
	Because the bias $\nbias_{n+1}^\policy$ of order $n+1$ is the bias of the Markov reward process $(-\nbias_n^\policy, \kernel^\policy)$, the proof naturally goes by induction on $n \ge 0$.
	For $n = 0$, since $\worstdiameter(\model) \ge \diameter(\kernel^\policy)$ by definition of $\worstdiameter(\model)$, the result follows readily from \Cref{lemma_variations_bias}, that bounds the variations of the $0$-th order bias, by setting $\alpha_0' (\model) := (12 + (16  + \abs{\states}) \diameter(\kernel_1)) \diameter(\kernel_1)$. 
	Supposing $n \ge 1$, the result follows by a straight-forward computation:
	\begin{align*}
		\norm*{
			\nbias_n^\policy - \hat{\nbias}_n^\policy
		}_\infty
		& \overset{\eqnum{1}}\le
		\parens*{
			12 + \parens*{16 + \abs{\states}} \diameter(\kernel_1)
		} \diameter(\kernel_1)
		\cdot 
		\mdpdistance
		\parens*{
			(-\nbias_{n-1}^\policy, \kernel^\policy), 
			(-\hat{\nbias}_{n-1}^\policy, \hat{\kernel}^\policy)
		}
		\\
		& \overset{\eqnum{2}}\le
		\parens*{
			12 + \parens*{16 + \abs{\states}} \diameter(\kernel_1)
		} \diameter(\kernel_1)
		\cdot 
		\alpha'_{n-1} (\model)
		\mdpdistance
		\parens{
			\model, \hat{\model}
		}
		\\
		& \equiv
		\parens*{
			12 + \parens*{16 + \abs{\states}} \diameter(\kernel_1)
		}^{n+1} \diameter(\kernel_1)^{n+1}
		\mdpdistance
		\parens{
			\model, \hat{\model}
		}
	\end{align*}
	where 
	\texteqnum{1} follows by \Cref{lemma_variations_bias}; 
	and
	\texteqnum{2} follows by induction. 
	Note that on the way, we show that $\alpha'_n (\model) = (12 + (16  + \abs{\states}) \diameter(\kernel_1))^{n+1} \diameter(\kernel_1)^{n+1}$.
	This concludes the proof. 
\end{proof}

    \section{CONSISTENCY AND PROOFS OF \texorpdfstring{SECTION~\ref{sect:consistency}}{SECTION 3}}\label{app:consistency}

\subsection{The \texorpdfstring{\texttt{HOPI}}{HOPI} Algorithm}

We start off by presenting \texttt{HOPI}, our algorithm capable of computing optimal policies of the desired order.
The pseudo-code is provided in \Cref{alg:HOPI}.

\begin{algorithm}
    \caption{Higher Order Policy Iteration}
    \label{alg:HOPI}
    \begin{algorithmic}[1]
        \STATE\textbf{Input:} Order $n$, slack $\varepsilon\geq 0$
        \STATE Initialize $k\gets 1$ and some arbitrary policy $\pi_k$
        \STATE $\triangleright$ \textit{Initialize a Bellman optimal policy of order $0$}
        \LOOP
            \STATE Compute $\gain^{\pi_k}$ and $\bias^{\pi_k}$
            \IF{$\gain^{\pi_k}$ is not constant}
                \STATE Select $\pi_{k+1}$ that is constant and such that $\gain^{\pi_{k+1}}(s)=\max_{s'\in \states} \gain^{\pi_k}(s')$
                \STATE $k\gets k+1$
            \ELSIF{there is $s\in\states$ with $\pi_k(s) \notin \soft_\varepsilon \argmax_a\{r(s,a)+p(s,a)\bias^{\pi_k}\}$}
                \STATE $\pi_{k+1}(s)\gets \soft_\varepsilon \argmax_a\{r(s,a)+p(s,a)\bias^{\pi_k}\}$ \hfill {\smaller (break ties arbitrarily)}
                \STATE $\pi_{k+1}(s') \gets \pi_k(s')$ for $s'\neq s$
                \STATE $k\gets k+1$
            \ELSE
                \STATE $\actions_{-2}\gets \actions$, $k_{-2}\gets k$
                \STATE \textbf{break}
            \ENDIF
        \ENDLOOP
        \STATE $\triangleright$ \textit{Progressively refine the output to Bellman optimal policies of order $n$}
        \FOR{$m=-1,0,\dots,n$}
            \LOOP
                \STATE Compute the bias vector $(\nbias_{m'}^{\policy_k})_{-1\leq m'\leq m+2}$
                \IF{there is $s\in\states$ with $\pi_k(s)\notin \soft_\varepsilon \argmax_{a\in \actions_{m-1}(s)} \{r_{m+1}(s,a)+p(s,a) \nbias^{\pi_k}_{m+1}\}$}
                    \STATE $\triangleright$ \textit{First Stage Policy Improvement}
                    \STATE $\pi_{k+1}(s)\gets \soft_\varepsilon \argmax_{a\in \actions_{m-1}(s)}\{r_{m+1}(s,a)+p(s,a)\nbias_{m+1}^{\pi_k}\}$ \hfill {\smaller (break ties arbitrarily)}
                    \STATE $\pi_{k+1}(s') \gets \pi_k(s')$ for $s'\neq s$
                    \STATE $k\gets k+1$
                \ELSE
                    \STATE $\triangleright$ \textit{Second Stage Policy Improvement}
                    \STATE $\actions_n^k (s) \gets \soft_\varepsilon \argmax_{a\in \actions_{m-1}(s)}\{r_{m+1}(s,a)+p(s,a)\nbias_{m+1}\}$
                    \IF{there is $s \in \states$ with $\pi_k(s) \notin \soft_\varepsilon \argmax_{a\in \actions_{m}^k(s)} \{r_{m+2}(s,a) +p(s,a)\nbias_{m+2}^{\pi_k}\}$}
                        \STATE $\pi_{k+1}(s) \gets \soft_\varepsilon \argmax_{a\in \actions_{m}^k(s)} \{r_{m+2}(s,a) +p(s,a)\nbias_{m+2}^{\pi_k}\}$\hfill {\smaller (break ties arbitrarily)}
                        \STATE $\pi_{k+1}(s') \gets \pi_k(s')$ for $s'\neq s$
                        \STATE $k\gets k+1$
                    \ELSE
                        \STATE $\actions_m(s) = \actions_m^k(s)$, $k_m \gets k$
                        \STATE \textbf{break}
                    \ENDIF
                \ENDIF
                \ENDLOOP
            \ENDFOR
        \STATE \textbf{return} $\actions_n$
    \end{algorithmic}
\end{algorithm}

The algorithm is essentially an ``epsilonized'' version of the high order Policy Iteration algorithm \cite[§10]{puterman_markov_1994}, that is, a version of Policy Iteration that takes noise into account during each of its iterations.
Because we will have to adapt the proof of Policy Iteration to a noisy setting, we provide a high level description of how the algorithm works.
The key idea of the algorithm is to compute the set of state-action pairs $\pairs_m$ that are optimal at order $m$, and to progressively increase $m$ to further refine the optimality order of the output policy until the desired order $n$.
That is, at stage $m \ge -2$, the goal of the algorithm is to construct a set of state-action pairs $\pairs_m \subseteq \pairs$ such that the associated set of policies
\begin{equation}
    \label{equation_hopi_1}
    \policies(\pairs_m)
    :=
    \braces*{
        \policy \in \policies
        :
        \forall \state \in \states,
        (\state, \policy(\state)) \in \pairs_m
    }
\end{equation}
satisfies $\optpolicies_{m+1} (\model) \subseteq \policies(\pairs_m) \subseteq \optpolicies_{m} (\model)$ (see \Cref{corollary_correctness_HOPI}).
To do so, the algorithm proceeds as follows.

At stage $m \ge -2$ and round $k$, the algorithm tries to refine the current policy $\policy_k$ with a two-staged process.
Write $\actions_\ell (\state) := \braces{\action: (\state, \action) \in \pairs_\ell}$ the right-projection of $\pairs_\ell$. Those are the actions that have been found to be optimal at order $\ell$.
\begin{itemize}
    \item (\textit{First Stage}) The algorithm looks at the Bellman optimal equations of order $m+1$ clipped to $\pairs_{m-1}$, the pairs that are known to be optimal at order $m-1$, and takes a soft argmax:
        \begin{equation}
            \label{equation_hopi_2}
            \actions_m^k (\state)
            :=
            \braces*{
                \action \in \actions_{m-1} (\state)
                :
                \reward_{m+1} (\state, \action)
                + \kernel(\state, \action) \nbias_{m+1}^{\policy_k}
                \ge
                \max_{\action' \in \actions_m (\state)}
                \braces*{
                    \reward_{m+1} (\state, \action')
                    + \kernel(\state, \action') \nbias_{m+1}^{\policy_k}
                }
                - \epsilon
            }
            .
        \end{equation}
        If the algorithm finds an improving state, i.e., $\state \in \states$ such that $\policy_k (\state) \notin \actions_m^k (\state)$, it improves the policy into $\policy_{k+1}$ and starts over to round $k+1$.
        Otherwise, it tries to further improve $\policy_k$ by moving to the second stage.

    \item (\textit{Second stage}) The algorithm looks at the Bellman optimal equations of order $m+2$ clipped to $\actions_m^k (\state)$, the $\epsilon$-optimal actions computed in the first stage, see \Cref{equation_hopi_2}, and takes a soft argmax:
        \begin{equation}
            \label{equation_hopi_3}
            \actions_{m+1}^k (\state)
            :=
            \braces*{
                \action \in \actions_{m}^k (\state)
                :
                \reward_{m+2} (\state, \action)
                + \kernel(\state, \action) \nbias_{m+2}^{\policy_k}
                \ge
                \max_{\action' \in \actions_m (\state)}
                \braces*{
                    \reward_{m+2} (\state, \action')
                    + \kernel(\state, \action') \nbias_{m+2}^{\policy_k}
                }
                - \epsilon
            }
            .
        \end{equation}
        Then it does the same as in the first stage:
        If the algorithm finds an improving state, i.e., $\state \in \states$ such that $\policy_k (\state) \notin \actions_{m+1}^k (\state)$, it improves the policy into $\policy_{k+1}$ and starts over to round $k+1$.
        Otherwise, $\actions_{m+1}^k$ is considered correct, and the algorithm sets
        \begin{equation}
            \pairs_m
            :=
            \bigcup_{\state \in \states}
            \braces*{\state}
            \times
            \actions_m^k (\state)
            .
        \end{equation}
        The algorithm sets $\policy_{k+1} \equiv \policy_k$, increases $m$ to $m+1$ and moves to the next round $k+1$, starting the new phase $m+1$, to compute $\pairs_{m+1}$.
\end{itemize}
On a side note, the case $m = -2$ is treated slightly differently than the others. Indeed, the usual criterion for Bellman optimality is that the policy must have constant gain (see Definition I.9 from \cite{boone_thesis_2024}), so this is what we use here.

The next few sections are dedicated to proving the correctness of \texttt{HOPI}.

\subsection{Links Between Optimalities and Gaps}

To begin with, note that we can work on the aperiodic transform of the underlying Markov decision process $\model \equiv (\pairs, \kernel, \reward)$, given by $\model' \equiv (\pairs, \kernel', \reward')$ where $\kernel'(\state'|\state, \action) := \frac 12 (\kernel(\state'|\state, \action) + \indicator{\state' = \state})$ and $\reward'(\state, \action) := \frac 12 \reward(\state, \action)$, see \cite[§8.5.4]{puterman_markov_1994}.
Indeed, we have $\nbias_n^{\policy} (\model') = 2^{n+2} \nbias_n^{\policy} (\model)$ (proof by induction on $n$), so that the optimality classes $\optpolicies_{n}$ and Bellman optimalities are unchanged.
Overall, we can assume freely that $\kernel(\state|\state, \action) \ge \frac 12$.
In particular, all policies are aperiodic.

The following lemma quantifies the differences between bias vectors of two policies with regards to their gaps. Depending on whether some state-action pairs are transient or recurrent under $\pi_2$, a different order of gap will appear.

\begin{lemma}
    \label{lem:gap-bias}
    Let $\model \equiv (\pairs, \kernel, \reward)$ be a Markov decision process such that $\kernel(\state|\state, \action) \ge \frac 12$.
    Let $\policy_1, \policy_2 \in \policies$ be two policies such that $\nbias_n^{\policy_1} = \nbias_n^{\policy_2}$ for $n \ge -1$.
    Then, for all $\state \in \states$ and $T \ge 1$, we have
    \begin{equation*}
        T \nbias_{n+1}^{\policy_2} (\state)
        + \nbias_{n+2}^{\policy_2} (\state)
        =
        T \nbias_{n+1}^{\policy_1} (\state)
        + \nbias_{n+2}^{\policy_1} (\state)
        -
        \EE_{\state}^{\policy_2} \brackets*{
            \nbias_{n+2}^{\policy_1} (\State_{T+1})
            +
            \sum_{t=1}^T
            \parens*{
                \gaps_{n+2}^{\policy_1}(\Pair_t)
                + (T - t) \gaps_{n+1}^{\policy_1}(\Pair_t)
            }
        }
        + \oh(1)
        .
    \end{equation*}

    For $n=-2$, we have for any $s \in \states$ that
    \begin{equation*}
        \gain^{\pi_2}(s)
        =
        g^{\pi_1}(s)
        - \lim_{T\rightarrow +\infty}
        \frac{1}{T}
        \bE^{\pi_2}_s \brackets*{
            \sum_{t=1}^T
            \parens*{
                \Delta_0^{\pi_1}(Z_{t})
                + (T-t) \Delta_{-1}^{\pi_1}(Z_t)
            }
        }
        .
    \end{equation*}
\end{lemma}
\begin{proof}
    Given a policy ${\pi_1}$, recall that the gap is defined by $\Delta_{n+1}^{\pi_1}(s,a) \coloneqq \nbias^{\pi_1}_{n+1}(s) + \nbias^{\pi_1}_n(s) - p(s,a)\nbias^{\pi_1}_{n+1} - r_{n+1}(s,a)$.
    Therefore, for any pair of policies ${\pi_1},{\pi_2}$, we have
    \begin{align*}
        r_{n+1}(s,a)
        - \nbias_n^{\pi_1}(s)
        & =
        \nbias_{n+1}^{\pi_1}(s)
        - p(s,a)\nbias_{n+1}^{\pi_1}
        - \Delta_{n+1}^{\pi_1}(s,a)
        \\
        \bE_s^{{\pi_2}} \brackets*{
            \sum_{t=1}^{T'-1}
            (r_{n+1}(Z_t) -\nbias_n^{\pi_1}(S_t))
        }
        & =
        \nbias_{n+1}^{\pi_1}(\state)
        - \bE^{{\pi_2}}_s \brackets*{
            \nbias_{n+1}^{\pi_1}(S_{T'})
        }
        - \bE^{{\pi_2}}_s \brackets*{
            \sum_{t=1}^{T'-1}
            \Delta_{n+1}^{\pi_1}(Z_t)
        }
        .
    \end{align*}
    Moreover, $r_{n+2}(s,a) - \nbias_{n+1}^{\pi_1}(s) = \nbias_{n+2}^{\pi_1}(s)-p(s,a)\nbias_{n+2}^{\pi_1} -\Delta_{n+2}^{\pi_1}(s,a)$, so that
    \begin{align}
        \nonumber
        &
        \bE_s^{{\pi_2}} \brackets*{
            \sum_{t=1}^{T'-1}
            (r_{n+1}(Z_t) - \nbias_n^{\pi_1}(S_t))
        }
        \\
        & \qquad =
        \label{eq:summingr-b}
        \nbias_{n+1}^{\pi_1} (\state)
        + \bE^{{\pi_2}}_s \brackets*{
            \nbias_{n+2}^{\pi_1}(S_{T'})
            - \nbias_{n+2}^{\pi_1}(S_{T'+1})
            - \Delta_{n+2}^{\pi_1}(Z_{T'})
            - r_{n+2}(Z_{T'})
        }
        - \bE^{{\pi_2}}_s \brackets*{
            \sum_{t=1}^{T'-1}
            \Delta_{n+1}^{\pi_1}(Z_t)
        }
        .
    \end{align}

    For $n> -2$, $r_{n+2}=0$.
    Summing over $T'=1,\dots,T$, we get
    \begin{align*}
        & \bE_s^{{\pi_2}} \brackets*{
            \sum_{T'=1}^T
            \sum_{t=1}^{T'-1}
            \parens*{
                r_{n+1}(Z_t)
                - \nbias_n^{\pi_1}(S_t)
            }
        }
        \\
        & \qquad =
        T \nbias_{n+1}^{\pi_1}(\state)
        + \nbias_{n+2}^{{\pi_1}}(s)
        - \bE^{{\pi_2}}_s \brackets*{
            \nbias_{n+2}^{\pi_1}(S_{T+1})
        }
        - \bE^{{\pi_2}}_s \brackets*{
            \sum_{T'=1}^T
            \parens*{
                \Delta_{n+2}^{\pi_1}(Z_{T'})
                +
                \sum_{t=1}^{T'-1}
                \Delta_{n+1}^{\pi_1}(Z_t)
            }
        }
        .
    \end{align*}

    Since $\nbias_n^{\pi_1}=\nbias_n^{\pi_2}$, the left hand term is equal to $\bE_s^{{\pi_2}} \left[ \sum_{T'=1}^T \sum_{t=1}^{T'-1} (r_{n+1}(Z_t) -\nbias_n^{\pi_2}(S_t))\right]$.
    Take the equation for $\pi_1=\pi_2$.
    Because for every state $s$, policy $\pi$ and integer $m$, we have $\gaps_m^{\pi}(s,\pi(s))=0$, it follows that
    \begin{align*}
        & T \nbias_{n+1}^{\policy_2} (\state)
        + \nbias_{n+2}^{\policy_2} (\state)
        \\
        & \qquad =
        T \nbias_{n+1}^{\policy_1} (\state)
        + \nbias_{n+2}^{\policy_1} (\state)
        + \EE_{\state}^{\policy_2} \brackets*{
            \nbias_{n+2}^{\policy_2} (\State_{T+1})
            - \nbias_{n+2}^{\policy_1} (\State_{T+1})
        }
        - \EE_{\state}^{\policy_2} \brackets*{
            \sum_{T'=1}^T
            \parens*{
                \gaps_{n+2}^{\policy_1} (\Pair_{T'})
                +
                \sum_{t=1}^{T'-1}
                \gaps_{n+1}^{\policy_1} (\Pair_{t})
            }
        }
        \\
        & \qquad \overset{\eqnum{1}}=
        T \nbias_{n+1}^{\policy_1} (\state)
        + \nbias_{n+2}^{\policy_1} (\state)
        + \EE_{\state}^{\policy_2} \brackets*{
            \nbias_{n+2}^{\policy_2} (\State_{T+1})
            - \nbias_{n+2}^{\policy_1} (\State_{T+1})
        }
        - \EE_{\state}^{\policy_2} \brackets*{
            \sum_{t=1}^T
            \parens*{
                \gaps_{n+2}^{\policy_1} (\Pair_{t})
                +
                (T - t)
                \gaps_{n+1}^{\policy_1} (\Pair_{t})
            }
        }
        \\
        & \qquad \overset{\eqnum{2}}=
        T \nbias_{n+1}^{\policy_1} (\state)
        + \nbias_{n+2}^{\policy_1} (\state)
        - \EE_{\state}^{\policy_2} \brackets*{
            \nbias_{n+2}^{\policy_1} (\State_{T+1})
        }
        - \EE_{\state}^{\policy_2} \brackets*{
            \sum_{t=1}^T
            \parens*{
                \gaps_{n+2}^{\policy_1} (\Pair_{t})
                +
                (T - t)
                \gaps_{n+1}^{\policy_1} (\Pair_{t})
            }
        }
        + \oh(1)
    \end{align*}
    where
    $\texteqnum{1}$ is obtained by reindexing the left-most sum and
    $\texteqnum{2}$ follows from $\EE_{\state}^{\policy_2}[\indicator{\State_{T+1} = \state'}] \to \imeasure^{\policy_2}(\state'|\state)$ where $\imeasure^{\policy_2}(-|\state)$ is a probability invariant measure of $\policy_2$ (by aperiodicity), combined with $\sum_{\state' \in \states} \imeasure^{\policy_2} (\state'|\state) \nbias_{n+2}^{\policy_2} (\state') = 0$.
    This concludes the proof for $n > -2$.

    For $n=-2$, \Cref{eq:summingr-b} becomes
    \begin{align*}
        0
        & =
        \gain^{\pi_1}(s)
        + \bE^{\pi_2}_s \brackets*{
            \bias^{\pi_1}(S_{T'})
            - \bias^{\pi_1}(S_{T'+1})
            - \Delta_0^{\pi_1}(Z_{T'})
            - \reward(Z_{T'})
            - \sum_{t=1}^{T'-1} \Delta_{-1}^{\pi_1}(Z_t)
        } \\
        & =
        T g^{\pi_1}(s)
        + \bias^{\pi_1}(s)
        - \bE^{\pi_2}_s \brackets*{
            \bias^{\pi_1}(S_{T+1})
        }
        - \bE^{\pi_2}_s \brackets*{
            \sum_{T'=1}^T
            \reward(Z_{T'})
        }
        - \bE^{\pi_2}_s \brackets*{
            \sum_{T'=1}^T \parens*{
                \Delta_0^{\pi_1}(Z_{T'})
                +
                \sum_{t=1}^{T'-1}
                \Delta_{-1}^{\pi_1}(Z_t)
            }
        }
    \end{align*}
    by summing over $T'=1,\dots,T$.
    This yields the result as $T\rightarrow +\infty$ for $\pi_1,\pi_2 = \pi,\pi'$.
\end{proof}

With this, we can prove the link between $n$-Bellman optimality and $(n-1)$-optimality, extending a result known for $n=0$.

\begin{proposition}
\label{proposition_bellman_to_optimal}
    Let $\model \equiv (\pairs, \kernel, \reward)$ be a communicating Markov decision process such that $\kernel(\state|\state, \action) \ge \frac 12$.
    For $m \ge -1$, a policy that is Bellman optimal at order $m$ is $(m-1)$-optimal.
\end{proposition}

\begin{proof}
    For $m = -1$, there is nothing to say because $\optpolicies_{-2} = \policies$.
    For $m = 0$, this is a well-known result, that a policy $\policy \in \policies$ satisfying the pair of optimality equations
    \begin{equation*}
        \begin{aligned}
            \forall \state \in \states,
            \quad
            \gain^{\policy} (\state)
            & =
            \max_{\action \in \actions(\state)}
            \braces*{
                \kernel(\state, \action)
                \gain^{\policy}
            }
            \\
            \forall \state \in \states,
            \quad
            \gain^{\policy} (\state)
            + \bias^{\policy} (\state)
            & =
            \max_{\action \in \actions(\state)}
            \braces*{
                \reward(\state, \action)
                +
                \kernel(\state, \action)
                \bias^{\policy}
            }
        \end{aligned}
    \end{equation*}
    has optimal gain. See for example Proposition I.4 from \cite{boone_thesis_2024}.
    We focus on the case $m \ge 1$.
    By induction, $\policy$ is $(m-2)$-optimal, i.e., $\nbias_k^* = \nbias_k^{\policy}$ for $k = -2, -1, \ldots, m-2$.
    In particular and taking $\optpolicy$ a Blackwell optimal policy, we have $\gaps^{\policy}_k = \gaps^{\optpolicy}_k$ for $k = -1, 0, \ldots, m-2$, so that $\gaps^{\policy}_k (\state, \optpolicy(\state)) = 0$ for all $\state \in \states$.
    Since $\policy$ is $(m-1)$-Bellman optimal, it follows that $\gaps_{m-1}^{\policy} (\state, \optpolicy(\state)) \ge 0$ for all $\state \in \states$.
    In all cases, invoking \Cref{lem:gap-bias} for $n \equiv m-1$, $\policy_1 = \policy$ and $\policy_2 = \optpolicy$, we find
    \begin{equation}
        \label{equation_temporary_1}
        \nbias_{m-1}^* (\state)
        =
        \nbias_{m-1}^\policy (\state)
        -
        \lim_{T \to \infty}
        \EE_{\state}^{\optpolicy} \brackets*{
            \frac 1T
            \sum_{t=1}^T
            \parens*{
                \gaps_{m}^{\optpolicy} (\Pair_t)
                + (T - t) \gaps_{m-1}^{\optpolicy} (\Pair_t)
            }
        }
        .
    \end{equation}
    The crucial additional observation is the following: Since $\policy$ is $m$-optimal, for every $\pair \in \pairs$ such that $\gaps_k^{\policy} (\pair) = 0$ for $k \le m-2$, either
    (i) $\gaps_{m-1}^{\policy} (\pair) > 0$; or
    (ii)~$\gaps_{m-1}^\policy (\pair) = 0$ and $\gaps_{m}^\policy (\pair) \ge 0$.
    Accordingly, we have (i) $\gaps_{m-1}^{\policy} (\Pair_t) > 0$ or (ii)~$\gaps_{m-1}^\policy (\Pair_t) = 0$ and $\gaps_{m}^{\policy} (\Pair_t) \ge 0$; $\Pr_{\state}^{\optpolicy}$-a.s.
    So,
    \begin{equation}
        \label{equation_temporary_2}
        \lim_{T \to \infty}
        \EE_{\state}^{\optpolicy} \brackets*{
            \frac 1T
            \sum_{t=1}^T
            \parens*{
                \gaps_{m}^{\optpolicy} (\Pair_t)
                + (T - t) \gaps_{m-1}^{\optpolicy} (\Pair_t)
            }
        }
        \ge
        0
        .
    \end{equation}
    Combining \Cref{equation_temporary_1,equation_temporary_2}, we conclude that $\nbias_{m-1}^* \le \nbias_{m-1}^\policy$.
    So $\policy$ is $(m-1)$-optimal.
\end{proof}

\subsection{\texorpdfstring{\texttt{HOPI}}{HOPI} Without Soft Argmax}

In this subsection, we use the previous results to prove the correctness of $\texttt{HOPI}(n,0)$ for any $n$, that is, of \texttt{HOPI} running without raw argmax rather than a soft argmax. 
The idea of the proof is the following.
First, we show that the two-phased policy improvement of \texttt{HOPI} improves the policy indeed, by showing that $(\nbias_m^{\policy_k})_{m \ge -1}$ is increasing with $k \ge 1$ for the lexicographic order, see \Cref{lem:polimpr}.  
Second, we show in \Cref{lem:induction} that the stopping rule of phase $m$ is correct: At the end of phase $m$, the policy $\policy_k$ is $(m+1)$-Bellman optimal. 
Combining the two, we deduce that \texttt{HOPI} runs in finite time, and runs a set of pairs $\pairs_n$ such that $\optpolicies_{n+1} \subseteq \policies(\pairs_n) \subseteq \optpolicies_n$. 

First of all, we build a policy improvement lemma: changing the first non-zero gap to be null improves the policy.

\begin{lemma}[Local Policy Improvement works]\label{lem:polimpr}
    Let $n \ge -1$. 
    Let $\pi$ and $\pi'$ be two policies, such that
    \begin{enumerate}[noitemsep]
        \item $\pi$ and $\pi'$ coincide on all states but one, denoted $s'$;
        \item For $m\leq n$, we have $\Delta_m^\pi(s',\pi'(s'))=0$;
        \item $\Delta_{n+1}^\pi(s',\pi'(s'))<0$.
    \end{enumerate}
    Then $\nbias_m^\pi = \nbias_m^{\pi'}$ for $m\leq n-1$, and $(\nbias^{\pi'}_n,\nbias^{\pi'}_{n+1}) > (\nbias^\pi_n,\nbias^\pi_{n+1})$ in lexicographic order.
\end{lemma}

\begin{proof}
    By induction, applying \Cref{lem:gap-bias} at orders $-2,\dots,n-2$ yields $\nbias^\pi_m = \nbias^{\pi'}_m$ for $m=-1,\dots,n-1$.

    Let us first assume that there exists a state $s$ such that $s'$ is recurrent under $\pi'$ starting from $s$, meaning the asymptotic visitation frequency $\mu_s^{\pi'}(s')>0$.
    Then for any such state $s$, \Cref{lem:gap-bias} yields
    \[\nbias_n^{\pi'}(s) = \nbias_n^{\pi}(s) - \Delta_{n+1}^\pi(s',\pi'(s')) \mu_s^{\pi'}(s') > \nbias_n^\pi(s)\] and for any other state $s$, $\nbias_n^{\pi'}(s)=\nbias_n^\pi(s)$, yielding the result.

    If there exists no such state, then $\nbias_n^{\pi'}=\nbias_n^{\pi}$, and we can apply \Cref{lem:gap-bias} at order $n$.
    \begin{align*} 
        \nbias^{\pi'}_{n+1} (s) 
        & = 
        \nbias^{\pi}_{n+1}(s) 
        - \lim_{T\rightarrow +\infty} 
        \bE^{\pi'}_s \brackets*{ 
            \sum_{T'=1}^T 
            \sum_{t=1}^{T'-1}
            \Delta_{n+1}^\pi(Z_t)
        }
    \end{align*}
    where $\lim_{T\rightarrow +\infty} \bE^{\pi'}_s [ \sum_{T'=1}^T \sum_{t=1}^{T'-1} \Delta_{n+1}^\pi(Z_t)] \leq 0$, and $\lim_{T\rightarrow +\infty} \bE^{\pi'}_{s'} [ \sum_{T'=1}^T \sum_{t=1}^{T'-1} \Delta_{n+1}^\pi(Z_t)] <0$ since starting in $s'$ guarantees that $s'$ is visited. Therefore, $\nbias_{n+1}^{\pi'} > \nbias_{n+1}^\pi$.
\end{proof}

This means that $(\pi_k)$ will have increasing bias vectors (in lexicographic order), which entails that algorithm $\texttt{HOPI}(n,0)$ necessarily stops (\Cref{corollary_hopi_stops}). 

\begin{corollary}
    \label{corollary_hopi_stops}
    \texttt{HOPI}$(n, 0)$ stops in finite time.
\end{corollary}

Even if the sequence of policies $(\policy_k)$ is forever improving until the algorithm stops, we have yet to prove that the output policy is indeed optimal at order $n+1$. 
We thus push \Cref{lem:polimpr} once step further, and show that once stage $n$ finishes with the policy $\pi_{k_n}$, we have found an optimal policy of order $n+1$, because no policy can best it.

\begin{lemma}[Phase $m$ stops with a $m$-Bellman optimal policy]
\label{lem:induction}
    For any $n\geq -1$, if $\Pi^\star_n\subseteq \policies(\pairs_{n-1}) \subseteq \Pi^\star_{n-1}$ and if $\pi_{k_{n-1}}$ is $(n+1)$-Bellman-optimal, then $\pi_{k_n}$ is $(n+2)$-Bellman optimal and $\Pi^\star_{n+1}\subseteq \policies(\pairs_{n}) \subseteq \Pi^\star_{n}$.
\end{lemma}

\begin{proof}
    Assume that $\Pi^\star_n\subseteq \policies(\pairs_{n-1}) \subseteq \Pi^\star_{n-1}$ and that $\pi_{k_{n-1}}$ is $(n+1)$-Bellman optimal.

    We start by proving that
    \begin{equation}
    \label{equation_induction_1}
        \policy_{k_n} \text{~is~$(n+2)$-Bellman optimal.}
    \end{equation}
    First of all, we claim that $\pi_k$ is $n$-optimal for all $k \ge k_{n-1}$. 
    This is shown by induction on $k \ge k_{n-1}$.
    By assumption, we know that $\policy_k$ is $n$-optimal for $k = k_{n-1}$.
    For the other inductive case, assume that $\pi_k$ is $n$-optimal. 
    For any state $\state$, $\pi_{k+1}$ satisfies $\Delta_{n+1}^{\pi_k}(s,\pi_{k+1}(s))\leq 0$ and $\Delta_n^{\pi_k}(s,\pi_{k+1}(s))=\Delta_n^{\pi_{k_{n-1}}}(s,\pi_{k+1}(s))=0$ by definition of $\pairs_{n-1}$ and by induction hypothesis.
    So, invoking \Cref{lem:gap-bias} up to order $n-1$, we see that $\policy_{k+1}$ has better bias $\nbias_m^{\policy_{k+1}}$ than $\policy_k$ for $m = -1, 0, \ldots, n$.
    As $\policy_k$ is $n$-optimal, we conclude that $\pi_{k+1}$ is $n$-optimal as well. 

    In particular, $\policy_{k_n}$ is $n$-optimal, so $(\nbias_m^{\pi_{k_n}})_{m\leq n} = (\nbias_m^{\pi_{k_{n-1}}})_{m\leq n}$, and $(\Delta_m^{\pi_{k_n}})_{m\leq n} = (\Delta_m^{\pi_{k_{n-1}}})_{m\leq n}$. 
    Because $\policy_{k_{n-1}}$ is $(n+1)$-Bellman optimal, it follows that $\policy_{k_n}$ is $(n+1)$-Bellman optimal. 
    To see that it is $(n+2)$-Bellman optimal, pick $\pair \in \pairs$ such that $\gaps_m^{\policy_{k_n}}(\pair) = 0$ for $m = -1, 0, \ldots, n+1$.
    Because $\policy_{k_n}$ and $\policy_{k_{n-1}}$ have the same gaps up to order $m = n$, we have $\pair \in \pairs_{n-1}$. 
    Then, by definition of $\pairs_n$ (see \Cref{alg:HOPI}), every $\pair' \in \pairs_{n-1}$ either satisfies 
    (1) $\gaps_{n+1}^{\policy_{k_n}}(\pair') \ge 0$, or 
    (2) $\gaps_{n+1}^{\policy_{k_n}}(\pair') = 0$ and $\gaps_{n+2}^{\policy_{k_n}}(\pair') \ge 0$. 
    By choice of $\pair$, (1) doesn't hold; so (2) does for $\pair' \equiv \pair$. 
    So by \Cref{definition_bellman_optimal_high_order}, $\policy_{k_n}$ is $(n+2)$-Bellman optimal, proving \eqref{equation_induction_1}.

    Now, we prove that $\optpolicies_{n+1} \subseteq \policies(\pairs_n)$. 
    Let $\pi \in \Pi_{n+1}^\star$.
    In particular, we have $\pi\in \Pi^\star_n$, so $\pi\in \policies(\pairs_{n-1})$ by assumption. 
    Recall that $(n+2)$-Bellman optimal policies are $(n+1)$-optimal (\Cref{proposition_bellman_to_optimal}), so that $\nbias_{n+1}^{\pi}=\nbias_{n+1}^{\pi_{k_n}}$.
    Therefore, for any state $\state \in \states$, $0=\Delta_{n+1}^{\pi}(s,\pi(s)) = \Delta_{n+1}^{\pi_{k_n}}(s,\pi(s))$.
    Hence $\pi \in \optpolicies(\pairs_n)$.

    Lastly, we prove that $\policies(\pairs_n) \subseteq \optpolicies_n$.
    Let $\pi \in \policies(\pairs_n)$.
    Since $\pairs_n \subseteq \pairs_{n-1}$, $\policy \in \policies(\pairs_{n-1})$. 
    So $\policy$ is $(n-1)$-optimal by assumption. 
    Furthermore, by defintion of $\pairs_n$, we have $\Delta_{n+1}^{\pi_{k_n}}(s,\pi(s))= 0$, $\Delta_{n}^{\pi_{k_n}}(s,\pi(s))= 0$ for every state $\state \in \states$.
    So, by invoking \Cref{lem:gap-bias} at order $n-1$, we obtain $\nbias_n^\pi \geq \nbias_n^{\pi_{k_n}} = \nbias_n^*$.
    Hence $\pi\in \Pi^\star_n$.
\end{proof}

\begin{corollary}[Correctness of \texttt{HOPI}]
\label{corollary_correctness_HOPI}
    For any $n\geq -2$, $\Pi^\star_{n+1} \subseteq \policies(\pairs_n) \subseteq \Pi^\star_n$.
\end{corollary}

\begin{proof}
    The proof goes by induction on $n = -2, -1, \ldots$. 

    At $n=-2$, for all state $\state \in \states$, we have $\pairs_{-2}(\state) = \pairs$, and therefore $\Pi = \policies(\pairs_{-2})$.
    Moreover, $\Pi^\opt_{-2}=\Pi$, so therefore we have $\Pi^\star_{-1}\subseteq \policies(\pairs_{-2}) \subseteq \Pi^\star_{-2}$.
    The fact that $\policy_{k-2}$ is 0-Bellman optimal is a consequence of well-known results \cite{puterman_markov_1994}, because the first loop of \texttt{HOPI}$(n, 0)$ consists in running the vanilla (first order) Policy Iteration algorithm, that returns a policy satisfying the Bellman equations; hence a 0-Bellman optimal policies.

    Therefore, we can propagate the property to larger $n$ using \Cref{lem:induction}. 
\end{proof}

With the previous result, we have the correctness of the $\texttt{HOPI}(n,0)$ algorithm. 

We now have to look into its approximate version, that replace the use of the raw $\argmax(-)$ operator by a soft argmax during the execution of the algorithm. 

\subsection{\texorpdfstring{\texttt{HOPI}}{HOPI} Running with Non-Trivial Soft Argmax}

The point of this section is to prove that, when both $\varepsilon > 0$ and $\mdpdistance(M,M')$ are small enough, then $\texttt{HOPI}(n,0,M) = \texttt{HOPI}(n,\varepsilon,M')$.
As our main result, we show that when $\epsilon$ and $\mdpdistance(\model, \model')$ are well-tuned with respect to each other, every step that $\texttt{HOPI}(n,\varepsilon,M')$ takes during its execution is the same as $\texttt{HOPI}(n,0,M)$.

\newcommand{\dgap}{\mathrm{dgap}}

Given a set $\mathcal{X} \subseteq \RR$, introduce
\begin{equation}
\label{equation_dgap}
    \dgap(\mathcal{X})
    :=
    \min \parens*{
        \braces*{
            \abs{x - y}
            :
            x \in \mathcal{X},
            y \in \mathcal{X},
            x \ne y
        }
    }
\end{equation}
the minimal gap between two distints elements of $\mathcal{X}$. 

\begin{lemma}\label{lemma_quantification_eps_appr}
    Let $M \equiv (\pairs,p,r)$ a communicating Markov decision process. Fix $n\geq 0$, a policy $\pi\in \Pi$, a state $s\in \states$ and a set of actions $\actions^\opt(s)\subseteq \actions(s)$. 
    Denote $\alpha_n(M) >0$ defined in \Cref{lemma_close_MDPs} and $\worstdiameter (\model)$ the worst-diameter of $\model$ (\Cref{definition_worst_diameter}).
    For all $\varepsilon>0$ and instance $\hat{M}\equiv(\pairs,\hat{p},\hat{r})$ that satisfy 
    \begin{equation}
    \label{equation_quantification_eps_appr}
        \mdpdistance(M,\hat{M}) 
        \leq 
        \min \braces*{
            \frac 1{D^\star(M)},
            \frac{\varepsilon}{2\alpha_n(M)},
            \frac{
                \dgap\parens*{
                    \braces*{
                        \gaps_n^\policy(\state, \action)
                        :
                        \action \in \actions^*(\state)
                    }
                }
                -\varepsilon
            }{2\alpha_n(M)}
        },
    \end{equation}
    we have
    \begin{equation*}
        \argmax_{a\in\actions^\star(s)} 
        \braces*{
            r_n(s,a) +p(s,a) \nbias_n^\pi
        } 
        = 
        \soft_\varepsilon 
        \argmax_{a\in\actions^\star(s)} 
        \braces*{
            \hat{r}_n(s,a) +\hat{p}(s,a) \hat{\nbias}_n^\pi 
        }
        .
    \end{equation*}
\end{lemma}

\begin{proof}
    We proceed by double inclusion.

    Let $\action \in \argmax_{a\in\actions^\star(s)} \left\{ r_n(s,a) +p(s,a) \nbias_n^\pi \right\}$. We have
    \begin{align*}
        \hat{r}_n(s,a) + \hat{p}(s,a)\hat{\nbias}^\pi_n 
        & \overset{\eqnum{1}}{\geq} 
        r_n(s,a) +p(s,a) \nbias_n^\pi -\alpha_n(M) \mdpdistance(M,\hat{M}) 
        \\
        & = 
        \max_{a' \in \actions^\star(s)} 
        \braces*{
            r_n(s,a')
            + p(s,a') \nbias_n^\pi
        } - \alpha_n(M) \mdpdistance(M,\hat{M})
        \\
        & \overset{\eqnum{1}}{\geq} 
        \max_{a'\in \actions^\star(s)}
        \braces*{
            \hat{r}_n(s,a') 
            + \hat{p}(s,a') \hat{\nbias}_n^\pi
        } 
        - 2\alpha_n(M) \mdpdistance(M,\hat{M})
    \end{align*}
    where both instances of \texteqnum{1} stem from \Cref{lemma_close_MDPs}, since $\mdpdistance(M,\hat{M}) \leq 1/D^\star(M)$ by \Cref{equation_quantification_eps_appr}. 
    We deduce that, since $2\alpha_n(M) \mdpdistance(M,\hat{M}) \leq\varepsilon$ by \Cref{equation_quantification_eps_appr} again, $a \in \soft_\varepsilon \argmax_{a\in \actions^\star(s)} \braces*{\hat{r}_n(s,a') + \hat{p}(s,a') \hat{\nbias}_n^\pi}$.

    Conversely, fix $a\in \soft_\varepsilon \argmax_{a\in \actions^\star(s)} \braces*{ \hat{r}_n(s,a') +\hat{p}(s,a') \hat{\nbias}_n^\pi }$. 
    Similarly, 
    \begin{equation*}
        r_n(s,a)
        + p_n(s,a)\nbias_n^\pi 
        \geq 
        \max_{a'\in \actions^\star(s)}
        \braces*{
            r_n(s,a') 
            + p(s,a') \nbias_n^\pi
        }
        - 2\alpha_n(M) \mdpdistance(M,\hat{M}) 
        - \varepsilon
    \end{equation*}
    Since $2\alpha_n(M) \mdpdistance(M,\hat{M}) \leq \dgap \parens*{\braces{\gaps_n^\policy (\state, \action'): \action' \in \actions^*(\state)}}$ by \Cref{equation_quantification_eps_appr} always, we conclude that $a \in \argmax_{a\in\actions^\star(s)} \left\{ r_n(s,a) +p(s,a) \nbias_n^\pi \right\}$, which concludes the proof.
\end{proof}

Strong of \Cref{lemma_quantification_eps_appr}, we may conclude that $\texttt{HOPI}(n, 0, \model) = \texttt{HOPI}(n, \epsilon, \model')$ under the right tuning of $\epsilon$ and $\mdpdistance(\model, \model')$. 

\begin{corollary}[Bissimulation]
\label{corollary_bissimulation}
    Let $\model \equiv (\pairs, \kernel, \reward)$ a communicating Markov decision process and fix $n \ge 0$. 
    Denote $\alpha_n(M) >0$ defined in \Cref{lemma_close_MDPs} and $\worstdiameter (\model)$ the worst-diameter of $\model$ (\Cref{definition_worst_diameter}).
    For all $\varepsilon>0$ and instance $\model'\equiv(\pairs,p',r')$ that satisfy 
    \begin{equation}
    \label{equation_bissimulation}
        \mdpdistance(M,\hat{M}) 
        \leq 
        \min \braces*{
            \frac 1{D^\star(M)},
            \frac{\varepsilon}{2\alpha_n(M)},
            \min_{\policy \in \policies}
            \min_{m \le n+2}
            \min_{\state \in \states}
            \frac{
                \dgap\parens*{
                    \braces*{
                        \gaps_m^\policy(\state, \action)
                        :
                        \action \in \actions(\state)
                    }
                }
                -\varepsilon
            }{2\alpha_m(M)}
        },
    \end{equation}
    we have $\texttt{\upshape HOPI}(n, 0, \model) = \texttt{\upshape HOPI}(n, \epsilon, \model')$. 
\end{corollary}

\begin{proof}
    This is essentially a bissimulation argument. 
    Let $(\policy_k)$ and $(\policy'_k)$ the respective sequences of policies computed by \texttt{HOPI}$(n, 0, \model)$ and \texttt{HOPI}$(n, \epsilon, \model')$.
    Let $\pairs_m, k_m$ (resp.~$\pairs'_m, k'_m$) the respective sequences of pairs and phases starting times of \texttt{HOPI}$(n, 0, \model)$ (resp.~of \texttt{HOPI}$(n, 0, \model')$). 
    By induction on $k \ge 1$, we show that $\policy_k = \policy'_k$, so that $\pairs_m = \pairs'_m$ and $k_m = k'_m$; In other words, both algorithms takes the same decisions at every point. 

    Indeed, assume that $\policy'_k = \policy_k$, and that both \texttt{HOPI}$(n, 0, \model)$ and \texttt{HOPI}$(n, \epsilon, \model')$ have the same pair mask $\pairs_0$ and are at the same state $m_0 \le n$. 
    Write $\actions_0 (\state) := \braces{\action \in \actions(\state): (\state, \action) \in \pairs_0}$.
    Because \Cref{equation_bissimulation} holds, by \Cref{lemma_quantification_eps_appr}, we have
    \begin{equation*}
        \forall \state \in \states,
        \quad
        \argmax_{\action \in \actions_0(\state)}
        \braces*{
            \reward_{m_0} (\state, \action)
            + \kernel(\state, \action) \nbias_{m_0}^{\policy_k}
        }
        =
        \soft_\epsilon
        \argmax_{\action \in \actions_0(\state)}
        \braces*{
            \reward'_{m_0} (\state, \action)
            + \kernel'(\state, \action) \nbias_{m_0}'^{\policy_k}
        }
        .
    \end{equation*}
    Therefore, the current policy will be improved at the first stage during the execution of \texttt{HOPI}$(n, 0, \model)$ if, and only if it is improved so for \texttt{HOPI}$(n, \epsilon, \model')$, and $\policy_{k+1} = \policy'_{k+1}$. 
    If the policy enters the second stage of policy improvement, we conclude that $\policy_k = \policy'_k$ with the same arguments. 
\end{proof}

\subsection{The Learning Variant of \texorpdfstring{\texttt{HOPI}}{HOPI}: \texorpdfstring{\texttt{HOPE}}{HOPE}}

Our algorithm \texttt{HOPE} (\Cref{alg:HOPE}) feeds the empirical model $\hat{\model}_t$ and a precision $\epsilon_t \equiv t^{-1/4}$ to \texttt{HOPI}, and outputs a policy generated by \texttt{HOPI}$(n, \epsilon_t, \hat{\model}_t)$ without modification. 
So, it remains to show that, at some point, \texttt{HOPE} does yield small enough $\varepsilon$ and a good enough approximation of the underlying instance $\model$ so that \Cref{lemma_quantification_eps_appr} can be applied. 
As \Cref{corollary_bissimulation} foreshadows, the empirical MDP must converge to the real one faster than $\varepsilon_t$ goes to 0. 

In other words, we prove \Cref{th_consistency} from the main text, restated below for convenience. 

\textbf{\Cref{th_consistency}.}
{
    \itshape
    For all $n \ge -1$, \texttt{\upshape HOPE}$(n)$ (\Cref{alg:HOPE}) is consistent for $\optpolicies_n$.
}

\begin{proof}[Proof of \Cref{th_consistency}]
    By definition of consistency, we want to show that, for each model $M$, $\Pr(\hat{\pi}_t\notin \Pi_n^\star(M))=\oh(1)$. For that, we fix a Markov decision process $M$ and, for any $\delta>0$, we will show the existence of a $t_\delta$ satisfying 
    \begin{equation*}
        \Pr(\exists t\geq t_\delta,\hat{\pi}_t \notin \Pi^\star_n(M))\leq \delta
        .
    \end{equation*}
    To achieve the above, the point is to invoke \Cref{corollary_bissimulation}, to state that \texttt{HOPI}$(n, 0, \model)$ and \texttt{HOPI}$(n, \epsilon_t, \hat{\model}_t)$ return the same thing, and conclude by correctness of \texttt{HOPI}$(n, 0, \model)$, see \Cref{corollary_correctness_HOPI}. 
    We want $t_\delta$ to be such that, $\varepsilon_{t_\delta} \leq \min_{m \le n+2} \frac{1}{2}\mathrm{dgap}(m; M)$, where
    \begin{equation}\label{eq:defdgapM}
        \dgap(m; \model)
        := 
        \min_{k \le m}
        \min_{\policy \in \policies}
        \dgap \parens*{
            \braces*{
                \gaps_k^\policy (\pair)
                ;
                \pair \in \pairs
            }
        }
    \end{equation}
    This is possible, since $\varepsilon_t \equiv t^{-1/4}$ is decreasing in t: it is satisfied for $t_\delta \geq t_0 := \left(\frac{2}{\min_{m\leq n+2} \mathrm{dgap}(m; \model)}\right)^4$.

    We also want $t_\delta$ to be such that $\mdpdistance(\model, \hat{\model}_t)$ is smaller than $\worstdiameter(M)^{-1}$ and $\varepsilon_t$ with high probability when $t\geq t_\delta$.

    For any $t_1\geq 1$,
    \begin{align*}
        & \Pr \parens*{
            \exists t\geq t_\delta, 
            \mdpdistance(\model, \hat{\model}_t)
            >
            \min \braces*{
                \frac 1{\worstdiameter(\model)},
                \varepsilon_t
            }
        }
        \\
        & \qquad \leq \Pr\left( \exists t\geq t_1, \exists \pair\in \pairs, \visits_t(\pair)<\beta t\right) 
        + \Pr \parens*{
            \substack{
                \displaystyle
                \exists t\geq t_1,
                \mdpdistance(\model, \hat{\model}_t)
                > \min\{D^\star(M)^{-1},\varepsilon_t\} 
                \\
                \displaystyle
                \mathrm{and~}
                \forall t\geq t_1, 
                \forall \pair \in \pairs, 
                \visits_t(\pair) \geq \beta t
            }
        }
        \\
        & \qquad \overset{\eqnum{1}}{\leq} 
        \exp(-\gamma t_1 +\lambda) 
        + \Pr \parens*{
            \substack{
                \displaystyle
                \exists t\geq t_1,
                \mdpdistance(\model, \hat{\model}_t)
                > \min\{D^\star(M)^{-1},\varepsilon_t\} 
                \\
                \displaystyle
                \mathrm{and~}
                \forall t\geq t_1, 
                \forall \pair \in \pairs, 
                \visits_t(\pair) \geq \beta t
            }
        }
    \end{align*}
    where $\texteqnum{1}$ comes from \Cref{lem_visits}. Suppose now that $t_1$ is such that, for $t\geq t_1$, 
    \begin{equation*}
        \min \braces*{
            \frac 1{\worstdiameter(\model)},
            \varepsilon_t
        } 
        \geq 
        \sqrt{
            \frac{|\states|}{\beta t}
            \log \parens*{
                \frac{8|\pairs|\sqrt{1+\beta t}}{\delta}
            }
        }
    \end{equation*}
    (there exists such a $t_1$ since $\varepsilon_t$ is of the form $\epsilon_t = \frac{1}{t^y}$ with $y<1/2$).
    Then,
    \begin{align*}
        & \Pr \parens*{
            \exists t\geq t_1,
            \norm{M-\hat{M}_t} > \min\{D^\star(M)^{-1},\varepsilon_t\}
        } \\
        &\hspace{2em} \leq 
        \exp(-\gamma t_1 +\lambda) 
        + 
        \Pr \parens*{
            \exists t\geq t_1,
            \norm{M-\hat{M}_t}
            > 
            \sqrt{
                \frac {\abs{\states}}{\beta t}
                {\log\left(\frac{8|\pairs|\sqrt{1+\beta t}}{\delta}\right)}
            } 
            \cap 
            \forall t\geq t_1, 
            \min(\visits_t) \geq \beta t
        } \\
        &\hspace{2em} \overset{\eqnum{1}}{\leq} 
        \exp(-\gamma t_1 +\lambda) 
        + 
        \Pr \parens*{
            \exists t\geq t_1,
            \norm{M-\hat{M}_t}
            > 
            \sqrt{
                \frac{\abs{\states}}{\beta t}
                {\log\left(  \frac{8|\states||\actions|\sqrt{1+\min(\visits_t)}}{\delta}\right)}
            } 
            \cap 
            \forall t\geq t_1, 
            \min(\visits_t) \geq \beta t
        } \\
        &\hspace{2em} \overset{\eqnum{2}}{\leq} 
        \exp(-\gamma t_1 +\lambda) 
        + \frac \delta2
    \end{align*} where $\texteqnum{1}$ is due to $x>0\mapsto \frac{\log y\sqrt{1+x}}{x}$ being decreasing when $y>1$, and $\texteqnum{2}$ invokes \Cref{cor:convergencespeed}.

    Defining $t_2 := \left\lceil \frac{1}{\gamma} \left( \ln \frac{\delta}{2} -\lambda\right)\right\rceil$, we thus have that when $t_\delta\geq \max\{t_1,t_2\}$, 
    \begin{equation*}
        \Pr \parens*{
            \exists t\geq t_\delta, 
            \mdpdistance(\model, \hat{\model}_t)
            >
            \min \braces*{
                \frac 1{\worstdiameter(\model)},
                \varepsilon_t
            }
        }
        \leq 
        \delta
        .
    \end{equation*}

    Let us put it all together now. With $t_\delta \geq \max\{t_0,t_1,t_2\}$,
    \begin{align*}
        &\Pr \parens*{
            \exists t\geq t_\delta,\hat{\pi}_t\notin\Pi^\star_n(M)
        } \\
        &\hspace{2em} \overset{\eqnum{1}}{\leq} 
        \Pr \parens*{
            \exists t\geq t_\delta, 
            \exists m\leq n+2, 
            \mdpdistance(\model, \hat{\model}_t)
            > 
            \min \braces*{
                \frac 1{\worstdiameter(\model)},
                \frac{\varepsilon_t}{2\alpha_m(M)},
                \frac{\mathrm{dgap}(m; M)-\varepsilon_t}{2\alpha_m(M)}
            }
        }\\
        &\hspace{2em} \leq 
        \Pr \parens*{
            \exists t\geq t_\delta, 
            \exists m\leq n+2, 
            \mdpdistance(\model, \hat{\model}_t)
            >
            \min \braces*{
                \frac 1{\worstdiameter(\model)},
                \frac{\varepsilon_t}{2\alpha_m(M)}
            }
            \cup \varepsilon_t > \frac{\mathrm{dgap}(m; M)}{2}
        } 
        \leq \delta
    \end{align*}
    where $\texteqnum{1}$ stems from \Cref{corollary_correctness_HOPI}, the correctness of $\texttt{HOPI}(n,0, \model)$, and \Cref{eq:defdgapM}.
\end{proof}

    \section{DEGENERATE MDPS AND PROOFS OF \texorpdfstring{SECTION~\ref{sect:degen}}{SECTION 4}}

\subsection{Proofs of \texorpdfstring{\Cref{section_nondegeneracy}}{Section 4}}

In this paragraph, we prove \Cref{proposition_non_degenerate_general} from the main text, of which we recall the statement below. 

\textbf{\Cref{proposition_non_degenerate_general}.}
{
    \itshape
    Let $\optpolicies : \models \to 2^\policies$ be a measurable optimality map. 
    If $\model \in \models$ is non-degenerate, then there is a policy $\policy \in \optpolicies(\model)$ that remains optimal in a neighbourhood of $\model$, i.e.,
    \begin{equation*}
        \exists \epsilon > 0,
        \quad
        \bigcap \braces*{
            \optpolicies(\model')
            :
            \mdpdistance(\model, \model') < \epsilon
        }
        \ne
        \emptyset
        .
    \end{equation*}
}

\begin{proof}[Proof of \Cref{proposition_non_degenerate_general}]
	Let $\model \in \models$ be a non-degenerate model. 
	By definition of non-degeneracy (\Cref{definition_non_degenerate}), there exists an identification algorithm $(\learner, \recommendation)$ that is consistent and such that, for all $\delta > 0$, there is a stopping time $\tau_\delta$ making $(\learner, \recommendation, \tau_\delta)$ $\delta$-PC and such that $\Pr^{\model, \learner}(\tau_\delta < \infty) = 1$. 
	Fix $\model' \in \models$ with $\distance_\epsilon (\model, \model') < \epsilon$ for some $\epsilon > 0$ to be specified later. 
	Given $\model'' \in \braces{\model, \model'}$, the probability operator $\Pr^{\model'', \learner}(-)$ on $\histories$ can be restricted to the sub-space $\histories_T$ of possible histories of length at time $T$; It is written $\Pr^{\model'', \learner}_T (-)$. 
	Now, for all $T \ge 1$, we have
	\begin{align*}
		& \Pr^{\model, \learner} \parens*{
			\recommendation_{\tau_\delta} \in \optpolicies(\model')
		}
		\\
		& \overset{\eqnum{1}}=
		\sup_{T \ge 1}
		\braces*{
			\Pr^{\model, \learner} \parens*{
				\recommendation_{\tau_\delta} \in \optpolicies(\model')
				\mathrm{~and~}
				\tau_\delta \le T
			}
		}
		\\
		& \overset{\eqnum{2}}=
		\sup_{T \ge 1}
		\braces*{
			\Pr^{\model, \learner}_T \parens*{
				\recommendation_{\tau_\delta} \in \optpolicies(\model')
				\mathrm{~and~}
				\tau_\delta \le T
			}
		}
		\\
		& \overset{\eqnum{3}}\ge
		\sup_{T \ge 1}
		\braces*{
			\Pr^{\model', \learner}_T \parens*{
				\recommendation_{\tau_\delta} \in \optpolicies(\model')
				\mathrm{~and~}
				\tau_\delta \le T
			}
			- \totalvariationdistance\parens*{
				\Pr^{\model, \learner}_T,
				\Pr^{\model', \learner}_T
			}
		}
		\\
		& \overset{\eqnum{4}}\ge
		\sup_{T \ge 1}
		\braces*{
			1 - \delta
			- \Pr^{\model', \learner} (\tau_\delta > T)
			- \totalvariationdistance\parens*{
				\Pr^{\model, \learner}_T,
				\Pr^{\model', \learner}_T
			}
		}
		\\
		& \overset{\eqnum{5}}\ge
		1 - \delta - 
		\inf_{T \ge 1}
		\braces*{
			\Pr^{\model, \learner} (\tau_\delta > T)
			+ 2 \totalvariationdistance\parens*{
				\Pr^{\model, \learner}_T,
				\Pr^{\model', \learner}_T
			}
		}
	\end{align*}
	where
	$\texteqnum{1}$ uses that $(\recommendation_{\tau_\delta} \in \optpolicies(\model'))$ is the increasing union of $(\recommendation_{\tau_\delta} \in \optpolicies(\model') \mathrm{~and~} \tau_\delta \le T)$;
	$\texteqnum{2}$ uses that the event $(\recommendation_{\tau_\delta} \in \optpolicies(\model') \mathrm{~and~} \tau_\delta \le T)$ is $\sigma(\History_T)$-measurable;
	$\texteqnum{3}$ follows by definition of the total-variation distance;
	$\texteqnum{4}$ follows by definition of the $\delta$-PC property; 
	$\texteqnum{5}$ follows by definition of the total-variation distance again. 
	
	Fix $\delta, x > 0$ with $\abs{\policies}(\delta + 2 x) < 1$, e.g., $\delta < \abs{\policies}^{-1}$ and $x < \frac 12 (\abs{\policies}^{-1} - \delta)$. 
	
	Because $\Pr^{\model, \learner}(\tau_\delta < \infty) = 1$, there exists $T \ge 0$ such that $\Pr^{\model, \learner}(\tau_\delta > T) \le x$.
	By induction on $T \ge 1$, we find that 
	\begin{equation*}
		\totalvariationdistance\parens*{
			\Pr^{\model, \learner}_T, \Pr^{\model', \learner}_T
		}
		\le 
		T \distance_\infty (\model, \model').
	\end{equation*}
	So, for $\distance_\infty (\model, \model') < \frac x{2T} =: \epsilon$, we get $\Pr^{\model, \learner}(\recommendation_{\tau_\delta} \in \optpolicies(\model')) \ge 1 - \delta - 2x$. 
	So
	\begin{equation}
		\label{equation_infinitesimal_halo_1}
		\Pr^{\model, \learner} \parens*{
			\recommendation_{\tau_\delta} \notin \optpolicies(\model') 
		}
		\le
		\delta + 2x.
	\end{equation}
	
	Now, let $\policies_\epsilon := \bigcap {\optpolicies(\model') : \distance_\infty (\model, \model') < \epsilon}$ be the set of policies that are universally optimal in the $\epsilon$-neighborhood of $\model$. 
	For every policy $\policy \notin \policies_\epsilon$, there is a model $\model'_\policy \in \models$ with $\distance_\infty (\model, \model'_\epsilon) < \epsilon$ that rejects $\policy$ in the sense that $\policy \notin \optpolicies(\model'_\epsilon)$.
	Following from this observation, we obtain
	\begin{align*}
		\Pr^{\model, \learner} \parens*{
			\recommendation_{\tau_\delta}
			\in 
			\policies_\epsilon
		}
		& =
		\Pr^{\model, \learner} \parens*{
			\forall \policy \notin \policies_\epsilon,
			\recommendation_{\tau_\delta} \in \optpolicies(\model'_\policy)
		}
		\\
		& =
		1 - \Pr^{\model, \learner} \parens*{
			\exists \policy \notin \policies_\epsilon,
			\recommendation_{\tau_\delta} \notin \optpolicies(\model'_\policy)
		}
		\\
		& \ge
		1 
		-
		\sum_{\policy \notin \policies_\epsilon}
		\Pr^{\model, \learner} \parens*{
			\recommendation_{\tau_\delta} \notin \optpolicies(\model'_\policy)
		}
		\\
		& \overset{\eqnum{1}}\ge
		1 - \abs{\policies \setminus \policies_\epsilon} (\delta + 2x)
		\ge
		1 - \abs{\policies} (\delta + 2x)
		\overset{\eqnum{2}}> 0
	\end{align*}
	where 
	$\texteqnum{1}$ follows from \eqref{equation_infinitesimal_halo_1}; and
	$\texteqnum{2}$ follows by definition of $\delta$ and $x$. 
	We conclude that $\policies_\epsilon \ne \emptyset$. 
\end{proof}

\subsection{Proofs of \texorpdfstring{\Cref{section_bellman_to_gain}}{Section 4.2}}

In this paragraph, we prove \Cref{proposition_shattering} and \Cref{lemma_isolating_bellman} from the main text (both re-stated below).
Recall that the point of \Cref{lemma_isolating_bellman} is only to serve in the proof of \Cref{proposition_shattering}.

\textbf{\Cref{lemma_isolating_bellman}.}
{
    \itshape
    Let $\model \equiv (\pairs, \rewarddistribution, \kernel)$ be a communicating instance.
    For every unichain Bellman optimal policy $\policy \in \bellmanpolicies(\model)$ and $\epsilon > 0$, there exists $\model' \equiv (\pairs, \reward', \kernel)$ with $\mdpdistance(\model, \model') < \epsilon$ for which
    \vspace{-1em} 
    \begin{enum}
        \item $\policy$ is the unique Bellman optimal policy;
        \item $\optgain(\model') = \optgain(\model)$; 
        \item $\optbias(\model') = \optbias(\model)$. 
    \end{enum}
}

\begin{proof}[Proof of \Cref{lemma_isolating_bellman}]
	Let $\policy \in \policies$ be a Bellman optimal policy of $\model$ and $\epsilon > 0$ be a precision parameter. 
	Consider the Markov decision process $\model_\epsilon \equiv (\pairs, \reward_\epsilon, \kernel)$ with mean reward function $\reward_\epsilon (\pair) := \epsilon + (1 - 2 \epsilon) \reward (\pair)$. 
	Accordingly, we have $\epsilon \le \reward_\epsilon (\pair) \le 1 - \epsilon$ for all $\pair \in \pairs$. 
	The policy $\policy$ is still Bellman optimal in $\model_\epsilon$ and we have $\distance_\infty (\model, \model_\epsilon) \le \epsilon$. 
	
	Consider the Markov decision process $\model'_\epsilon \equiv (\pairs, \reward'_\epsilon, \kernel)$ with mean reward function given by 
	\begin{equation*}
		\reward'_\epsilon (\state, \action)
		:=
		\reward_\epsilon (\state, \action)
		- \epsilon \indicator{\action \ne \policy(\state)}
		.
	\end{equation*}
	Note that $\gainof{\policy}(\model'_\epsilon) = \gainof{\policy}(\model_\epsilon)$ and $\biasof{\policy}(\model'_\epsilon) = \biasof{\policy}(\model_\epsilon)$ because the reward and kernel functions of $\policy$ are the same in $\model_\epsilon$ and $\model'_\epsilon$. 
	Now, for $\action \ne \policy(\state)$, we have 
	\begin{align*}
		\reward'_\epsilon (\state, \action)
		+ \kernel(\state, \action) \biasof{\policy}(\model'_\epsilon)
		& \overset{\eqnum{1}}=
		\reward_\epsilon (\state, \action)
		+ \kernel (\state, \action) \biasof{\policy}(\model_\epsilon)
		- \epsilon
		\\
		& \overset{\eqnum{2}}\le
		\gainof{\policy}(\state; \model_\epsilon)
		+ \biasof{\policy}(\state; \model_\epsilon)
		- \epsilon
		\overset{\eqnum{3}}=
		\gainof{\policy}(\model'_\epsilon)
		+ \biasof{\policy}(\model'_\epsilon)
		- \epsilon
	\end{align*}
	where
	$\texteqnum{1}$ follows by definition of $\reward'_\epsilon$ and $\biasof{\policy}(\model'_\epsilon) = \biasof{\policy}(\model_\epsilon)$;
	$\texteqnum{2}$ holds because $\policy$ is bias optimal in $\model_\epsilon$; and
	$\texteqnum{3}$ follows from $\gainof{\policy}(\model'_\epsilon) = \gainof{\policy}(\model_\epsilon)$ and $\biasof{\policy}(\model'_\epsilon) = \biasof{\policy}(\model_\epsilon)$. 
	Together with $\vecspan{\gainof{\policy}(\model'_\epsilon)} = \vecspan{\gainof{\policy}(\model_\epsilon)} = 0$, we conclude that $\policy$ is the unique Bellman optimal policy of $\model'_\epsilon$. 
	By triangular inequality, we further have $\distance_\infty(\model, \model'_\epsilon) \le 2 \epsilon$, concluding the proof. 
\end{proof}

\Cref{lemma_isolating_bellman} will be combined to the following well-known policy improvement result. 

\begin{lemma}[First order policy improvement; \cite{puterman_markov_1994}]
	\label{lemma_first_order_policy_improvement}
	Fix $\model \equiv (\pairs, \reward, \kernel)$ a Markov decision process.
	Let $\policy \in \policies$ satisfying $\vecspan{\gainof{\policy}} = 0$.
	If $\policy' \in \policies$ is such that $\gapsof{\policy}(\state, \policy'(\state)) \le 0$ for all $\state \in \states$, then $(\gainof{\policy}, \biasof{\policy}) \lexicographicorder (\gainof{\policy'}, \biasof{\policy'})$ where $\lexicographicorder$ is the lexicographic order. 
\end{lemma}

\begin{proof}
    The proof is essentially the same as the one of \Cref{lem:polimpr}, by changing $<$ to $\le$. 
\end{proof}

Now, we can prove \Cref{proposition_shattering}.

\textbf{\Cref{proposition_shattering}.}
{
    \itshape
    Let $\model$ be a communicating instance.
    For every unichain Bellman optimal policy $\policy \in \bellmanpolicies(\model)$ and precision $\epsilon > 0$, there exists $\model' \in \models$ with $\mdpdistance(\model, \model') < \epsilon$ for which $\policy$ is the unique gain optimal policy, i.e., $\optpolicies_{-1}(\model') = \braces{\policy}$. 
}

\begin{proof}[Proof of \Cref{proposition_shattering}]
	Let $\policy$ be a Bellman optimal policy of $\model$ and fix a precision $\epsilon > 0$.
	By \Cref{lemma_isolating_bellman}, there exists $\model' \equiv (\pairs, \reward', \kernel)$ such that $\distance_\infty (\model, \model') \le \epsilon$ where 
	(1)~$\policy$ is the unique Bellman optimal policy,
	(2)~$\optgain(\model') = \optgain(\model)$ and 
	(3)~$\optbias(\model') = \biasof{\policy}(\model)$.
	Consider the model $\model'' \equiv (\pairs, \reward'', \kernel'')$ with reward and kernel functions given by
	\begin{align*}
		\reward'' (\state, \action)
		& =
		\reward' (\state, \action)
		+ (\kernel'' (\state, \action) - \kernel' (\state, \action)) \biasof{\policy}(\model),
		\\
		\kernel'' (\state'|\state, \action)
		& =
		(1 - \epsilon) \kernel' (\state, \action)
		+ \frac \epsilon{\abs{\actions(\state)}} 
		.
	\end{align*}
	Note that $\norm{\kernel''(\pair) - \kernel(\pair)}_1 \le \abs{\states} \epsilon$, so that $\norm{\reward'' - \reward'}_\infty \le \frac 12 \vecspan{\biasof{\policy}(\model)} \abs{\states} \epsilon$ by \Cref{lemma_l1_deviation_span}.
	Further note that $\model''$ is uniformly ergodic, since all transition probabilities are positive. 
	
	Consider another policy $\policy' \in \policies$. 
	Since $\policy'$ is ergodic in $\model''$, there exists $c(\policy') > 0$ such that every state is visited at least $c(\policy')$ in average in $\model''$. 
	For $\state \in \states$, we have
	\begin{align*}
		\gainof{\policy'}(\state; \model'')
		& =
		\lim_{T \to \infty}
		\frac 1T 
		\EE_{\state}^{\policy', \model''} \brackets*{
			\sum_{t=1}^T 
			\reward'' (\Pair_t)
		}
		\\
		& \overset{\eqnum{1}}=
		\lim_{T \to \infty}
		\frac 1T
		\EE_{\state}^{\policy', \model''} \brackets*{
			\sum_{t=1}^T
			\parens*{
				\reward'(\Pair_t)
				+ \parens*{
					\kernel'(\Pair_t) - \kernel''(\Pair_t)
				} \biasof{\policy}(\model')
			}
		}
		\\
		& \overset{\eqnum{2}}=
		\lim_{T \to \infty}
		\frac 1T
		\EE_{\state}^{\policy', \model''} \brackets*{
			\sum_{t=1}^T
			\parens*{
				\gainof{\policy}(\State_t; \model')
				+ \parens*{
					\unit_{\State_t} - \kernel''(\Pair_t)
				} \biasof{\policy}(\model')
				- \gapsof{\policy} (\Pair_t; \model')
			}
		}
		\\
		& \overset{\eqnum{3}}=
		\gainof{\policy}(\state; \model')
		+
		\lim_{T \to \infty}
		\frac 1T
		\EE_{\state}^{\policy', \model''} \brackets*{
			\biasof{\policy}(\state; \model')
			- 
			\biasof{\policy}(\State_{T+1}; \model')
			-
			\sum_{t=1}^T
			\gapsof{\policy} (\Pair_t; \model')
		}
		\\
		& \overset{\eqnum{4}}\le
		\gainof{\policy}(\state; \model')
		- c(\policy')
		\max_{\state \in \states}
		\ogaps(\state, \policy'(\state); \model')
	\end{align*}
	where
	$\texteqnum{1}$ unfolds the definition of $\reward''$ and uses that $\biasof{\policy}(\model) = \biasof{\policy}(\model')$; 
	$\texteqnum{2}$ follows by definition of the gaps of $\policy$, i.e., $\gapsof{\policy}(\state, \action) = \gainof{\policy}(\state) + \biasof{\policy}(\state) - \reward(\state, \action) - \kernel(\state, \action) \biasof{\policy}$;
	$\texteqnum{3}$ rewrites the telescopic sum $\EE[\sum_{t=1}^T (\unit_{\State_t} - \kernel''(\Pair_t)) \biasof{\policy}(\model')]$; and 
	$\texteqnum{4}$ follows by definition of $c(\policy)$ and uses that $\gapsof{\policy}(\model') = \ogaps(\model')$ that holds because $\policy$ is Bellman optimal in $\model'$. 
	
	From $\texteqnum{3}$, note that for $\policy = \policy'$, we obtain
	\begin{equation*}
		\gainof{\policy}(\model'') = \gainof{\policy}(\model')
	\end{equation*}
	since pairs played by $\policy$ satisfy $\gapsof{\policy}(\pair; \model') = 0$ by construction of $\model'$. 
	
	Let $\dmin(\ogaps(\model')) := \min \braces*{\ogaps(\pair; \model'): \pair \in \pairs ~\mathrm{and}~\ogaps(\pair; \model') > 0}$ the minimum positive gap of $\model'$. 
	Because $\policy$ is the unique Bellman optimal policy in $\model'$, for every $\policy' \ne \policy$, there is $\state \in \states$ such that $\gapsof{\policy}(\state, \policy'(\state); \model') > 0$---Indeed, if, ad absurdum $\gapsof{\policy}(\state, \policy'(\state); \model') = 0$ for all $\state \in \states$, then because $\policy$ is the unique Bellman optimal policy, it is bias optimal.
	So, by \Cref{lemma_first_order_policy_improvement}, we see that $\policy'$ is bias optimal as well, so is Bellman optimal in particular; A contraction. 
	
	Continuing, for every policy $\policy' \ne \policy$, we get 
	\begin{equation*}
		\gainof{\policy'}(\model'')
		\le
		\gainof{\policy}(\model)
		- \min_{\policy''} c (\policy'')
		\dmin(\ogaps)
		<
		\gainof{\policy}(\model')
		=
		\gainof{\policy}(\model'')
		.
	\end{equation*}
	Accordingly, $\policy$ is the unique gain optimal policy of $\model''$.
	Furthermore, the distance between $\model''$ and $\model$ is $\OH(\epsilon)$. 
	This concludes the proof. 
\end{proof}

	\section{STOPPING OF \texttt{HOPE} AND PROOFS OF \texorpdfstring{SECTION~\ref{sect:stopping}}{SECTION 5}}

\subsection{Proof of \Cref{lemma_deciding_unique_bellman}}

In this paragraph, we prove \Cref{lemma_deciding_unique_bellman}, restated below for convenience.

\textbf{\Cref{lemma_deciding_unique_bellman}.}
{
    \itshape
    Let $\model$ be a communicating instance. 
    Then

    \vspace{-0.33em} 
    \begin{enum}
        \item
            If $\model$ has a unique Bellman optimal policy, then this policy is unichain.
        \item 
            Conversely, if there exists a unichain policy $\policy \in \policies$ such that $\gainof{\policy}(\state) + \biasof{\policy}(\state) > \reward(\state, \action) + \kernel(\state, \action) \biasof{\policy}$ for all $\action \ne \policy(\state)$ and $\state \in \states$, then $\policy$ is the unique Bellman optimal policy of $\model$. 
    \end{enum}
}

\begin{proof}[Proof of \Cref{lemma_deciding_unique_bellman}]
	We prove both directions separately. 
	
	The first assertion (1) is proved by contradiction.
	Assume that $\model$ has a unique Bellman optimal policy $\policy$ that is multi-chain.
	Let $\states_1$ and $\states_2$ be two disjoint recurrent classes of states of $\policy$. 
	For $i \in \braces{1, 2}$ and $\epsilon > 0$, consider the Markov decision processes $\model^\epsilon_i \equiv (\pairs, \kernel, \reward^i_\epsilon)$ with reward functions given by $\reward^\epsilon_i (\state, \action) = \reward(\state, \action) - \epsilon \indicator{\state \notin \states_i}$. 
	Then, for $\model^\epsilon_1$ for instance, we see that $\policy$ does not satisfy $\vecspan{\gainof{\policy}(\model_1^\epsilon)} = 0$---Yet $\model_1^\epsilon$ is communicating, so $\vecspan{\optgain(\model_1^\epsilon)} = 0$. 
	So $\policy$ is sub-optimal in $\model_1^\epsilon$. 
	Therefore, $\model_1^\epsilon$ must admit a Bellman optimal policy $\policy_1^\epsilon$ that is not $\policy$.
	Because $\policies$ is finite, we see that there exists $\epsilon_n \to 0$ and $\policy' \ne \policy$ such that $\policy' \in \bellmanpolicies_0 (\model_1^{\epsilon_n})$ for all $n \ge 0$. 
	Accordingly, for all $n \ge 1$ and $\state \in \states$,
	\begin{equation}
		\label{equation_structure_unique_bellman_1}
		\gainof{\policy'}(\state; \model_1^{\epsilon_n})
		+
		\biasof{\policy'}(\state; \model_1^{\epsilon_n})
		=
		\max_{\action \in \actions(\state)}
		\braces*{
			\reward_1^{\epsilon_n} (\state, \action)
			+
			\kernel(\state, \action) \biasof{\policy'}(\model_1^{\epsilon_n})
		}
	\end{equation}
	with $\vecspan{\gainof{\policy'}(\model_1^{\epsilon_n})} = 0$.
	Because the gain $\gainof{\policy'}$ and bias functions $\biasof{\policy'}$ of $\policy'$ are linear functions of the mean reward vector, so by making $n \to \infty$, we see that \eqref{equation_structure_unique_bellman_1} and $\vecspan{\gainof{\policy'}} = 0$ are satisfied for $\model$ as well.
	In other words, $\policy'$ is Bellman optimal in $\model$; A contradiction. 
	
	The second assertion (2) follows from a standard argument that is reminiscent of the analysis of Policy Iteration.
	Let $\policy \in \policies$ be a unichain policy such that $\gapsof{\policy}(\state, \action) := \gainof{\policy}(\state) + \biasof{\policy}(\state) - \reward(\state, \action) - \kernel(\state, \action) \biasof{\policy} > 0$ for all $\action \ne \policy(\state)$. 
	Because it is unichain, we have $\vecspan{\gainof{\policy}} = 0$ so $\policy$ is Bellman optimal in $\model$. 
	
	Let $\policy'$ a Bellman optimal policy.
	We show that $\policy = \policy'$. 
	We have
	\begin{equation}
		\label{equation_structure_unique_bellman_2}
		\EE_{\state}^{\policy'} \brackets*{
			\Reward_1 + \ldots + \Reward_T
		}
		=
		T \gainof{\policy}(\state)
		+ \EE_{\state}^{\policy'} \brackets*{
			\biasof{\policy}(\state) - \biasof{\policy}(\State_{T+1})
		}
		- \EE_{\state}^{\policy'} \brackets*{
			\sum_{t=1}^T
			\gaps^{\policy}(\Pair_t)
		}
	\end{equation}
	and since $\gainof{\policy}(\state) = \gainof{\policy'}(\state)$, we conclude that every state $\state \in \states$ such that $\gapsof{\policy}(\state, \policy'(\state)) > 0$ must be transient under $\policy'$. 
	So, $\policy'$ must coincide with $\policy$ on its recurrent states. 
	In particular, every recurrent class of $\policy'$ is a recurrent class of $\policy$.
	Because $\policy$ is unichain, we conclude that $\policy'$ is unichain as well as both policies have the unique recurrent class of states $\states_0$. 
	In particular, $\biasof{\policy} = \biasof{\policy'}$ in $\states_0$. 
	Now,
	\begin{equation}
		\label{equation_structure_unique_bellman_3}
		\EE_{\state}^{\policy'} \brackets*{
			\Reward_1 + \ldots + \Reward_T
		}
		=
		T \gainof{\policy}(\state)
		+ \EE_{\state}^{\policy'} \brackets*{
			\biasof{\policy'}(\state) - \biasof{\policy'}(\State_{T+1})
		}
		.
	\end{equation}
	Combining \Cref{equation_structure_unique_bellman_2,equation_structure_unique_bellman_3}, we find that for all $\state \in \states$, we have
	\begin{align*}
		\biasof{\policy'}(\state)
		- \EE_{\state}^{\policy'} \brackets*{\biasof{\policy'}(\State_{T+1})}
		& =
		\biasof{\policy}(\state)
		- \EE_{\state}^{\policy'} \brackets*{\biasof{\policy}(\State_{T+1})}
		- \EE_{\state}^{\policy'} \brackets*{\sum_{t=1}^T \gapsof{\policy}(\Pair_t)}
		\\
		& =
		\biasof{\policy}(\state)
		- \EE_{\state}^{\policy'} \brackets*{\biasof{\policy'}(\State_{T+1})}
		+ \EE_{\state}^{\policy'} \brackets*{
			\biasof{\policy'}(\State_{T+1})
			- \biasof{\policy}(\State_{T+1})
		}
		- \EE_{\state}^{\policy'} \brackets*{\sum_{t=1}^T \gapsof{\policy}(\Pair_t)}
		\\
		& \overset{\eqnum{1}}\le
		\biasof{\policy}(\state)
		- \EE_{\state}^{\policy'} \brackets*{\biasof{\policy'}(\State_{T+1})}
		+ \norm*{\biasof{\policy'} - \biasof{\policy}}_\infty
		\Pr_{\state}^{\policy'} \parens*{\State_{T+1} \notin \states_0}
		- \EE_{\state}^{\policy'} \brackets*{\sum_{t=1}^T \gapsof{\policy}(\Pair_t)}
		\\
		& =
		\biasof{\policy}(\state)
		- \EE_{\state}^{\policy'} \brackets*{\biasof{\policy'}(\State_{T+1})}
		- \EE_{\state}^{\policy'} \brackets*{\sum_{t=1}^T \gapsof{\policy}(\Pair_t)}
		+ \oh(1)
	\end{align*}
	where $\texteqnum{1}$ follows from $\biasof{\policy} = \biasof{\policy'}(\state)$.
	We conclude that $\biasof{\policy'}(\state) \le \biasof{\policy}(\state)$ with strict inequality if $\policy(\state) \ne \policy'(\state)$.
	Yet, symmetrically, we find that $\biasof{\policy} \le \biasof{\policy'}$. 
	So $\biasof{\policy} = \biasof{\policy'}$ and $\policy = \policy'$ necessarily. 
\end{proof}

\subsection{Proof of \Cref{section_stopping_rule}}

In this paragraph, we prove \Cref{lemma_neighborhood_non_degenerate,proposition_hope_stopping_rule} from the main text.

\textbf{\Cref{lemma_neighborhood_non_degenerate}.}
{
    \itshape
    Let $\model \equiv (\pairs, \rewarddistribution, \kernel)$ be a communicating instance with unique Bellman optimal policy $\optpolicy$. 
    For every $\model' \equiv (\pairs, \rewarddistribution', \kernel')$ such that $\support(\kernel') \supseteq \support(\kernel)$, if 
    \begin{equation*}
        \mdpdistance(\model, \model')
        <
        \beta(\model)
    \end{equation*}
    then $\optpolicy$ is the unique Bellman optimal policy of $\model'$.
}

\begin{proof}[Proof of \Cref{lemma_neighborhood_non_degenerate}]
	Since $\support(\kernel') \supseteq \support(\kernel)$, $\optpolicy$ is unichain in $\model'$ as well. 
	So, by \Cref{lemma_deciding_unique_bellman}, we see that $\bellmanpolicies_0 (\model') = \braces{\optpolicy}$ if, and only if $\support \parens{\gapsof{\optpolicy}(\model')} = \support \parens{\gapsof{\optpolicy}(\model)}$. 
	Because $\gapsof{\optpolicy}(\state, \action; \model) > 0$ for all $\action \ne \optpolicy(\state)$, we deduce that $\bellmanpolicies_0 (\model') = \braces{\optpolicy}$ holds in particular when
	\begin{equation*}
		\norm*{
			\gapsof{\optpolicy}(\model')
			- \gapsof{\optpolicy}(\model)
		}_\infty
		< \dmin \parens*{\ogaps(\model)}.
	\end{equation*}
	We conclude with the explicit bound on the variations of the gap function of \Cref{lemma_unichain_gap_variations}.
\end{proof}

\textbf{\Cref{proposition_hope_stopping_rule}.}
{
    \itshape
    Upon running with the stopping rule 
    \begin{equation*}
        \tau_\delta
        :=
        \inf \braces*{
            t \ge 1
            :
            \xi_\delta (t)
            \le 
            \beta (\hat{\model}_t)
            \mathrm{~and~}
            \bellmanpolicies(\hat{\model}_t) = \braces{\recommendation_t}
        }
    \end{equation*}
    where $\xi_\delta(t)$ and $\beta(t)$ are respectively given by \Cref{equation_threshold_confidence,equation_threshold_non_degenerate}, \texttt{HOPE} is $\delta$-PC for $\optpolicies_n$, whatever $n \ge 1$.
    Moreover, for every instance $\model$ with unique Bellman optimal policy, we have $\Pr^{\model, \texttt{HOPE}} (\tau_\delta < \infty) = 1$. 
}

\begin{proof}[Proof of \Cref{proposition_hope_stopping_rule}]
	By \Cref{cor:convergencespeed}, we know that $\Pr^{\model, \learner}(\exists t \ge 1, \distance_\infty (\hat{\model}_t, \model) \ge \xi_\delta (t)) \le \delta$.
	Combined with \Cref{lemma_neighborhood_non_degenerate}, we know that if $\optpolicy_t = \recommendation_t$ is the unique Bellman optimal policy of $\hat{\model}_t$, then every model $\model' \gg \hat{\model}_t$ that satisfies $\distance_\infty (\hat{\model}_t, \model') \le \xi_\delta (t)$ further satisfies $\bellmanpolicies_0 (\model') = \braces{\optpolicy_t}$. 
	We deduce that 
	\begin{equation*}
		\Pr^{\model, \learner} \parens*{
			\tau_\delta < \infty
			\mathrm{~and~}
			\optpolicies(\model) \ne \braces*{\recommendation_{\tau_\delta}}
		}
		\le
		\delta
		.
	\end{equation*}
	In particular, the algorithm is $\delta$-PC. 
	
	To conclude, we show that $\Pr^{\model, \learner}(\tau_\delta < \infty) = 1$.
	Pick $\model$ a model with a unique Bellman optimal policy $\optpolicy$.
	By consistency, we have
	\begin{equation}
		\label{equation_delta_pc_1}
		\Pr^{\model, \learner} \parens*{
			\recommendation_t = \optpolicy
		}
		=
		1 + \oh(1)
	\end{equation}
	when $t \to \infty$. 
	As actions are played uniformly, we have $\visits_t (\pair) \to \infty$ a.s.
	So $\hat{\model}_t \to \model$ a.s.~and together with the deviations results of \Cref{lemma_unichain_gap_variations,lemma_unichain_gain_variations,lemma_unichain_bias_variations,lemma_unichain_reaching_time_variations}, we see that $\beta (\hat{M}_t) \to \beta(M)$ a.s. 
	Combined with \Cref{equation_delta_pc_1}, we obtain in particular
	\begin{equation}
		\label{equation_delta_pc_2}
		\Pr^{\model, \learner} \parens*{
			\recommendation_t = \optpolicy_t = \optpolicy
			\mathrm{~and~}
			\beta(\hat{M}_t) \ge \frac 12 \beta(M)
		} 
		=
		1 + \oh(1)
	\end{equation}
	when $t \to \infty$. 
	Moreover, we know that there exists $\alpha > 0$ such that $\Pr(\lim_{t \to \infty} \frac 1t \visits_t (\pair) > \alpha) = 1$. 
	It follows that $\xi_\delta (t) \to 0$ a.s.
	In tandem with \Cref{equation_delta_pc_2}, we conclude that $\Pr^{\model, \learner} (\recommendation_t = \optpolicy_t = \optpolicy, \beta(\hat{M}_t) \ge \frac 12 \beta(M) \mathrm{~and~} \xi_\delta (t) \le \frac 12 \beta(M)) = 1 + \oh(1)$. 
	Hence $\Pr^{\model, \learner}(\tau_\delta < \infty) = 1$. 
	\end{proof}

\end{document}